\documentclass[11pt,a4paper]{article}
\usepackage[hyperref]{acl}
\usepackage{times}
\usepackage{latexsym}

\usepackage{microtype}
\usepackage{graphicx}
\usepackage{comment}
\usepackage{booktabs}
\usepackage{dirtytalk}
\usepackage{cleveref}

\DeclareUnicodeCharacter{0307}{\c{c}}

\title{Repairing the Cracked Foundation: \\ A Survey of Obstacles in Evaluation Practices for Generated Text}

\author{Sebastian Gehrmann \\ \\ \\ \And 
        Elizabeth Clark \vspace{0.5em} \\ 
    Google Research \\ 
    New York, NY \\
  \texttt{\{gehrmann, eaclark, tsellam\}@google.com} \\ \And
  Thibault Sellam \\ \\ \\
  }

\date{}

\begin{document}
\maketitle
\begin{abstract}
Evaluation practices in natural language generation (NLG) have many known flaws, but improved evaluation approaches are rarely widely adopted. 
This issue has become more urgent, since neural NLG models have improved to the point where they can often no longer be distinguished based on the surface-level features that older metrics rely on. 
This paper surveys the issues with human and automatic model evaluations and with commonly used datasets in NLG that have been pointed out over the past 20 years. 
We summarize, categorize, and discuss how researchers have been addressing these issues and what their findings mean for the current state of model evaluations. 
Building on those insights, we lay out a long-term vision for NLG evaluation and propose concrete steps for researchers to improve their evaluation processes.
Finally, we analyze 66 NLG papers from recent NLP conferences in how well they already follow these suggestions and identify which areas require more drastic changes to the status quo.
\looseness=-1 
\end{abstract}

\section{Introduction}


There are many issues with the evaluation of models that generate natural language. For example, datasets are often constructed in a way that prevents measuring tail effects of robustness, and they almost exclusively cover English. Most automated metrics measure only similarity between model output and references instead of fine-grained quality aspects (and even that poorly). Human evaluations have a high variance and, due to insufficient documentation, rarely produce replicable results.

These issues have become more urgent as the nature of models that generate language has changed without significant changes to how they are being evaluated.
While evaluation methods can capture surface-level improvements in text generated by state-of-the-art models (such as increased fluency) to some extent, they are ill-suited to detect issues with the content of model outputs, for example if they are not attributable to input information.
These ineffective evaluations lead to overestimates of model capabilities. Deeper analyses uncover that popular models fail even at simple tasks by taking shortcuts, overfitting, hallucinating, and not being in accordance with their communicative goals. 

Identifying these shortcomings, many recent papers critique evaluation techniques or propose new ones. But almost none of the suggestions are followed or new techniques used. There is an incentive mismatch between conducting high-quality evaluations and publishing new models or modeling techniques. 
While general-purpose evaluation techniques could lower the barrier of entry for incorporating evaluation advances into model development, their development requires resources that are hard to come by, including model outputs on validation and test sets or large quantities of human assessments of such outputs. Moreover, some issues, like the refinement of datasets, require iterative processes where many researchers collaborate. All this leads to a circular dependency where evaluations of generation models can be improved only if generation models use better evaluations. 

\begin{figure*}[ht]
    \centering
    \includegraphics[width=\textwidth]{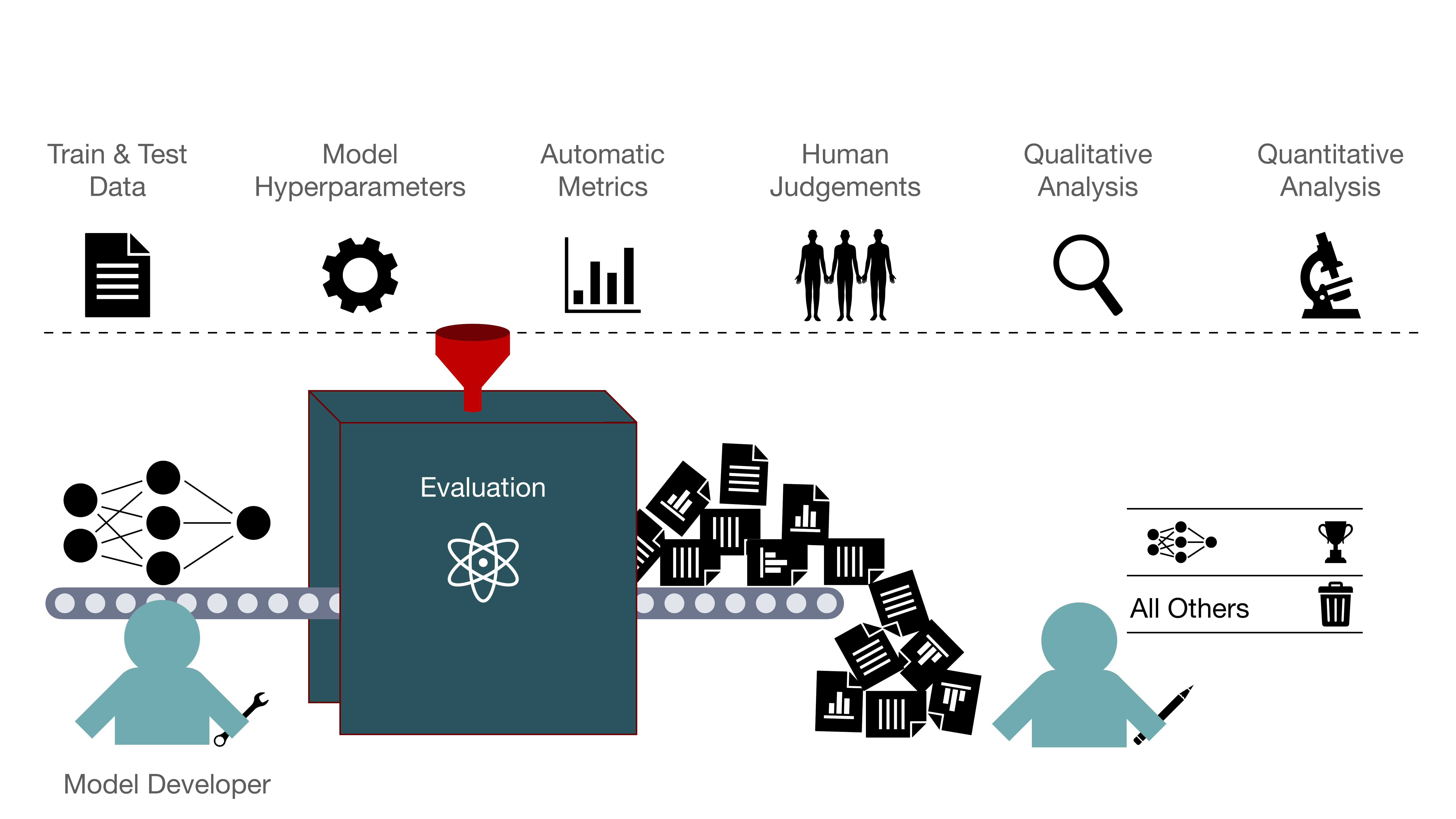}
    \caption{Even though the evaluation pipeline of a model is complex, with many steps and potential missteps that get ``funneled'' into the final results, it is often seen as a black box with the purpose of generating numbers that demonstrate superiority over competing approaches. We argue that more attention should be paid to the evaluation process and that the reporting of evaluation results should focus on the characteristics and limitations of a model.}
    \label{fig:teaser}
\end{figure*}

We find that there is a systemic difference between selecting the best model and characterizing how good this model really is. Current evaluation techniques focus on the first, while the second is required to detect crucial issues.
More emphasis needs to be put on measuring and reporting model limitations, rather than focusing on producing the highest performance numbers. To that end, this paper surveys analyses and critiques of evaluation approaches (\cref{sec:automatic,sec:human}) and of commonly used NLG datasets (\cref{sec:data}). Drawing on their insights, we describe how researchers developing modeling techniques can help to improve and subsequently benefit from better evaluations with methods available today (\cref{sec:recommendations}). Expanding on existing work on model documentation and formal evaluation processes~\citep{DBLP:conf/fat/MitchellWZBVHSR19,ribeiro-etal-2020-beyond}, we propose releasing evaluation reports which focus on demonstrating NLG model shortcomings using evaluation suites. These reports should apply a complementary set of automatic metrics, include rigorous human evaluations, and be accompanied by data releases that allow for re-analysis with improved metrics. 

In an analysis of 66 recent EMNLP, INLG, and ACL papers along 29 dimensions related to our suggestions (\cref{sec:survey}), we find that the first steps toward an improved evaluation are already frequently taken at an average rate of 27\%. The analysis uncovers the dimensions that require more drastic changes in the NLG community. For example, 84\% of papers already report results on multiple datasets and more than 28\% point out issues in them, but we found only a single paper that contributed to the dataset documentation, leaving future researchers to re-identify those issues.  
We further highlight typical unsupported claims and a need for more consistent data release practices.
Following the suggestions and results, we discuss how incorporating the suggestions can improve evaluation research, how the suggestions differ from similar ones made for NLU, and how better metrics can benefit model development itself~(\cref{sec:discussion}).

\section{Background}
\label{sec:background}

While ``natural language generation'' used to have a very narrow scope,\footnote{\citet{reiter1997building} define NLG as the process of producing text from structured data and thus, text-to-text or unconditional generation tasks would not count as NLG.} today it is used broadly to refer to the production of natural language in any context, and \textit{NLG tasks} include summarization, machine translation, paraphrasing, and story generation. 
For the purpose of this survey, we follow this broader definition, but focus on \textbf{conditional} generation tasks.
We define conditional NLG tasks as those in which a machine learning model can be trained to maximize a conditional probability $p(y|x)$ where $y$ is natural language and $x$ is an input that can be structured data or natural language and which provides information about what should be generated.\footnote{We omit multimodal tasks like image captioning or speech-to-text, as well as those with non-textual output like sign language or audio from the scope of this survey since those tasks require vastly different evaluation processes.}
The evaluation of conditionally generated text typically involves a comparison to the input and/or a reference text, neither of which is available in an unconditional generation setting.
The scope of this survey thus includes tasks such as machine translation, summarization, and data-to-text generation, but excludes language modeling.

In addition, we require in-scope NLG tasks to have an explicit \textbf{communicative goal}, which needs to be expressed while also planning the content and structure of the text and actualizing it in fluent and error-free language~\citep{gehrmann2020human}.\footnote{This requirement excludes most question-answering tasks since they require generating spans or otherwise non-fluent sequences of text.} 
All these aspects need to be captured in the NLG evaluation, making it much more challenging than evaluating other NLP tasks. 
For an introduction to NLG beyond this survey, we point readers to the overview by \citet{DBLP:journals/jair/GattK18} for a deeper discussion of NLG tasks, and to the survey by \citet{DBLP:journals/corr/abs-2006-14799} of the evaluation approaches and statistical methods that are discussed in Sections~\ref{sec:automatic}-\ref{sec:human}.


Evaluation approaches for generated text have traditionally been categorized as intrinsic or extrinsic~\citep{jones1995evaluating}. Intrinsic approaches evaluate a text by itself, whereas extrinsic approaches measure how it affects people performing a given task. Intrinsic evaluations include assessments by human ratings and by automatic metrics which have gained popularity with the advent of statistical NLG~\citep{langkilde-knight-1998-generation-exploits}, which led to the standardization of tasks. While some work exists that aims to standardize extrinsic evaluations~\citep[e.g.,][]{mani-etal-1999-tipster,gehrmann-etal-2019-improving}, the design space is much larger. As a result, intrinsic approaches dominate academic publications; \citet{gkatzia-mahamood-2015-snapshot} found that about 75\% of published NLG systems rely on intrinsic evaluations with the fraction increasing.\footnote{Informally surveying recent *CL papers suggests a number of 90\% or higher.}
Since we survey widely used approaches, we mostly cover intrinsic evaluations, but stress the importance of task-specific extrinsic evaluations. 

As pointed out by \citet{DBLP:journals/coling/ReiterB09}, the evaluation meta-evaluations we draw on are most commonly conducted on summarization and machine translation (MT), but that there is an implicit assumption that findings translate to other tasks. To avoid this issue, we note the task for each study, but, due to a lack of prior findings, are not able to cover every NLG task. Taking a cautious approach, we make the worst-case assumption that modes of failure likely transfer across tasks.

\section{Challenges of Automatic Evaluation}
\label{sec:automatic}

In this section, we provide an overview of common design principles of (intrinsic) automatic evaluation metrics, how these metrics are typically evaluated, what issues are being found, and how newly introduced metrics may overcome these issues in the future. Since not all evaluation strategies are being applied to all metrics and not all metrics are applied to all possible generation tasks, we can only provide an incomplete insight into the \textit{metric}$\times$\textit{task}$\times$\textit{evaluation method} space. Since there currently exists no ``perfect'' metric, we will not conclude with explicit metric recommendations but rather try and extract successful metric design principles alongside a family of evaluations that together may provide a more complete characterization of a model's performance.

\subsection{The Status Quo}
\label{sec:automatic-metrics}

Almost all commonly used generation metrics are reference-based: a system output $o$ is compared to one or multiple human-produced references, $\{r_1, \ldots, r_n\}$. System outputs that are more similar to the references are deemed better. However, there have been many strategies to measure the similarity. The most popular evaluation metrics, BLEU~\citep{papineni-etal-2002-bleu} and \textsc{ROUGE}~\citep{lin-2004-ROUGE}, along many others, measure the \textbf{lexical overlap} between $o$ and $r$ in terms of precision and recall of n-grams. Variants and parameters control tokenization, stemming, or balancing of precision and recall. 
With the advent of deep learning, metrics were introduced that measure the \textbf{distributional similarity} instead that rely on various ways to measure the distance between two distributed token and sequence representations. Notable examples from this class of metrics are the word mover distance~\citep{DBLP:conf/icml/KusnerSKW15}, which relies on non-contextual word embeddings, and \textsc{BERT-Score}~\citep{zhang20bertscore}, which aggregates cosine distances between represented tokens in a sequence, among others~\citep[][inter alia]{zhao-etal-2019-moverscore,clark-etal-2019-sentence,kane-etal-2020-nubia,Colombo21automatic}. 
A related class of automatic evaluation are \textbf{statistical approaches}, which focus on the distributions, rather than representations, produced by a model. \citet{saggion-etal-2010-multilingual} first demonstrated that distributional differences between references and model-outputs can be used as a scoring mechanism. \citet{gehrmann-etal-2019-gltr} showed that these differences exist even for large pretrained models, a fact that was used by \citet{zellers19defending} to train a classifier that detects generated text. \citet{hashimoto-etal-2019-unifying} used the same foundation to combine human and automatic evaluation in capturing the trade-off between sampling diverse outputs and achieving the highest possible quality. 
\citet{pillutla21mauve} expand on these insights and a framework by \citet{DBLP:conf/aistats/DjolongaLCBBG20} to compare the human- and model-distributions by measuring the extent to which they diverge. 
An alternative approach by \citet{thompson-post-2020-automatic} uses the probabilities of each model-generated token under a paraphrasing model that uses the human reference as input. 

Utilizing existing corpora of human quality judgments of generated text, \textbf{learned metrics} are classifiers that emulate these judgments. Some metrics move beyond reference-based evaluation and instead provide quality estimation scores between an input $i$ and output $o$. The first metric of this kind was \textsc{CLASSY}, a logistic regression model for summarization evaluation~\citep{DBLP:journals/algorithms/RankelCS12}. Newer metrics rely on pretrained models, are trained on more human ratings, and introduce initialization and and pretraining schemes~\citep[][inter alia]{sellam-etal-2020-bleurt,rei-etal-2020-comet,pu-etal-2021-learning, wegmann-nguyen-2021-capture}, or focus on specific aspects like the faithfulness of generated text~\citep[e.g.,][]{kryscinski-etal-2020-evaluating, aralikatte-etal-2021-focus}. Many of these metrics rely on artificially introduced errors, but \citet{cao-etal-2020-factual} find that moving from artificial to real error detection is challenging, an issue that \citet{Zeng21Gradient} aim to address by using adversarial examples instead. 

The metrics mentioned so far operate on text directly, but there has also been a long history of \textbf{metrics that generate and use intermediate structures}. These include accuracy of parse trees~\citep{bangalore-etal-2000-evaluation}, overlap between \say{basic elements}~\citep{hovy2005evaluating},\footnote{\textsc{ROUGE} is a special case of this where basic elements are fixed size n-grams, but other basic element metrics like PARENT~\citep{dhingra-etal-2019-handling} only focus on content words.} automatically constructed content units~\citep{tauchmann-mieskes-2020-language} using the Pyramid framework by \citet{nenkova-passonneau-2004-evaluating}, dependency parses~\citep{pratapa21evaluating}, or sequence alignment~\citep{deng21compression}. A special case of intermediate structures that recently gained popularity are \textbf{question-answering metrics} that assess information-equivalence. Similar to the faithfulness classifiers above, these aim to measure whether generated text contains the same information as a source or reference. Instantiations of these metrics may blank out entities~\citep{eyal-etal-2019-question,xie-etal-2021-factual-consistency,scialom-etal-2019-answers}, or fully generate questions~\citep[][inter alia]{DBLP:conf/aaai/00010W018,wang-etal-2020-asking,durmus-etal-2020-feqa,scialom21questeval,rebuffel21dataquesteval,honovich21evaluating,DBLP:journals/tacl/DeutschBR21}.

This overview already points to the first issue with the state of metrics research: the metrics listed above, except those targeting machine translation, are designed to work only on English. A notable exception is a study by \citet{Briakou21Evaluating} which assesses different learned metrics for formality transfer and uses multilingual pre-trained models such as XLM-R~\cite{conneau-etal-2020-unsupervised}. While automatic metrics are well-studied, the barrier of entry to developing non-English models is growing.

\subsection{Similarity to References is a Red Herring}
\label{sec:metrics-likeness}

Many automatic metrics rely on the assumption that NLG systems outputs that are more similar to the reference(s) are better, a property commonly referred to as ``human-likeness'' in the NLG literature (see, e.g., \citet{belz-gatt-2008-intrinsic}). While the ability to reproduce a reference text sounds like natural evidence of success, relying entirely on it for evaluation is misleading---a caveat pointed out by many evaluation researchers.
For instance, \citet{belz-gatt-2008-intrinsic} investigate the correlation between lexical overlap metrics (such as \textsc{BLEU} and \textsc{ROUGE}) and various measures of success in a Referring Expression Generation context. They find that ``a system’s ability to produce human-like outputs may be completely unrelated to its effect on human task-performance.'' 

One reason for this discrepancy is that similarity-based evaluations reward surface similarity at the expense of meaning and may be ``fooled'' by similar-looking, yet semantically different, outputs.
NLG tasks have an extensive output space which cannot be captured through a limited number of references and, a comparison to references becomes less reliable the more ``open-ended'' a task is. For that reason, \textsc{ROUGE} underperforms on non-extractive summaries~\citep{dorr-etal-2005-methodology}. The problem is especially poignant when the references themselves are flawed. As \citet{dhingra-etal-2019-handling} show, using \textsc{BLEU} and \textsc{ROUGE} is problematic with many table-to-text datasets, because there is a mismatch between the information conveyed by the reference texts and that of the input table. As a result, model outputs that contain similar unsupported information are rewarded by the metric. Similarly, \citet{freitag-etal-2020-bleu} show that \textsc{BLEU}, \textsc{METEOR}, and \textsc{BERTScore} may fail to reward good translations when the reference text contains artifacts such as ``translationese''.

One may wonder whether the problem still exists with learnt or embedding-based metrics, since a more flexible notion of similarity should enable metrics to be less reliant on surface-level features or text artifacts in references. However, this argument assumes that the set of reference appropriately covers the target domain, and that the metric is flexible enough to ``generalize'' from an incomplete set of examples. The current empirical evidence for this is negative ---in section~\ref{sec:automatic-audits} we will present several studies that show that even current metrics break down with simple adversarial examples~\cite{Sai21Perturbation, Kaster21Global}.


\paragraph{How to Interpret Similarity-Based Metrics?}


If similarity to the reference is a flawed proxy for quality, what \textit{do} automatic metrics tell us? This question can be investigated empirically by measuring the correlation between metric scores and human annotations. In a survey of such studies by \citet{reiter-2018-structured} focused on \textsc{BLEU}, he concludes that it is useful as a diagnostic tool during the development of MT systems, but not for other tasks and that is should not be used at the segment level. More recently, \citet{kocmi-etal-2021-ship} assess how well automatic metrics compute pairwise rankings for MT systems, and recommend using a combination of overlap-based and pretraining-based metrics, confirming the previous findings that metrics may be used to rank MT models at the system-level.

Several authors have tried to introduce finer-grained quality criteria, and attempted to understand which quality dimensions are captured by automatic metrics that measure the similarity to references. In most cases, there is inconclusive evidence. For instance, \citet{reiter-belz-2009-investigation} find that these metrics may approximate \textbf{language quality}, although with only weak evidence, and that they do not measure \textbf{content quality} at all. In contrast, \citet{DBLP:conf/cicling/StentMS05} evaluate metrics on restructured sentences, showing that lexical-overlap based metrics do measure similarity in meaning, but fail at measuring syntactic correctness. The inconsistency between studies and use cases suggests that overlap-based metrics likely measure neither, which is confirmed by later studies.

In a more recent study, \citet{kryscinski-etal-2019-neural} 5-way annotated system outputs on 100 samples from the test set of the CNN-Dailymail summarization corpus~\citep[CNNDM,][]{DBLP:conf/nips/HermannKGEKSB15,nallapati-etal-2016-abstractive} along two measures of content quality (relevance of the content, and faithfulness) and two of linguistic quality (on the sentence- and summary-level) using raters from Mechanical Turk. Consistent with previous findings, they find that \textsc{ROUGE} does not significantly correlate with either of them. Extending the annotations by three expert judgments per data point and extending the analysis to more metrics, \citet{Fabbri21summeval} find similarly low correlations without significant performance improvements of distributional over lexical similarity metrics.
Comparing correlations of these metrics across shared tasks from the Text Analysis Conferences (TAC) and CNN/DM and using a different annotation scheme, \citet{bhandari-etal-2020-evaluating} corroborate the very low segment-level correlations and also find that that no distributional metric outperforms \textsc{ROUGE}. 
Reanalyzing the data and addressing issues in the statistical tests, \citet{10.1162/tacl_a_00417} come to the same conclusion about \textsc{ROUGE}, but note the insights should be carefully assessed since the data selection strategy for annotations, coupled with large confidence intervals, can lead to false results.
Beyond summarization, \citet{novikova-etal-2017-need} note similarly poor segment-level correlations for data-to-text datasets.

All this shows that it is unclear what the results of embedding-based and lexical metrics represent, and it is questionable whether the numbers they produce can be trusted outside a few cases such as MT systems ranking. To better understand their limitations and opportunities, we need large-scale corpora of high-quality human annotations, which do not yet exist for most NLG tasks.

\paragraph{The Myth of the Single Reliable Number}

If human-likeness should not be used as proxy measure for quality of generated text, what should be used instead? 
Analyzing DUC 2004 data~\citep{over2004introduction}, where human raters annotated the language quality and the coverage of a summary, i.e., how well it covered the meaning of the source, \citet{graham-2015-evaluating} found that there was almost no correlation between the two measures. However, language quality was a precondition for achieving high coverage, leading to a complex relationship between the two. The lack of correlation between language and content quality was also noted by \citet{pitler-etal-2010-automatic} who find correlations between \textit{some} evaluation categories. These insights, combined with the lack of strong correlations, suggests that a single number, as produced by almost all automatic metrics, cannot fully characterize an NLG system. Similar points are made by \citet{deutsch-roth-2021-understanding} who show that many similarity metrics capture the overlap in topics between two summaries much better than the overlap in their information.

\paragraph{Faithfulness is Not Single Dimensional Either}

An aspect of quality mentioned above and which permeates all of NLG is \textbf{faithfulness}, and much recent work has focused on this aspect for abstractive summarization. \citet{maynez-etal-2020-faithfulness} state that a model is not faithful if it hallucinates, that is, it adds information that is not present in the source document. They define multiple categories of hallucinations: \textit{Intrinsic} hallucinations misrepresent facts in the input, for example turning a \say{former London mayoral candidate} into a \say{former London mayor}. \textit{Extrinsic} hallucinations ignore the input altogether, for example generating \say{President Sara} in the example above. Not all hallucinations are problematic---an extrinsic hallucination can be \textit{factual}, and may, in fact, be desirable depending on the use case. For system evaluation, it is therefore important to be able to discern between hallucinations of different types, which cannot be done by producing a single number.

\citeauthor{maynez-etal-2020-faithfulness} demonstrate that similarity metrics fail to measure faithfulness. The same failure is observed by \citet{pagnoni-etal-2021-understanding} who introduce and collect annotations for an alternative typology of factual errors which involves fine-grained categories such as \emph{Coreference Error} and \emph{Out of Article Error}. In an alternative approach to measuring correlations with human judgments, \citet{Gabriel21GoFigure} inject factual errors in reference summaries, and checks whether system rankings produced by metrics correlate with the ``level of factuality'' of the transformed sentences, among other properties like a metric's value range and generalization. They also identify that standard evaluation metrics (e.g., \textsc{ROUGE-L} and \textsc{ROUGE-1}) oftentimes fail at capturing factuality, but identify question-answering metrics as promising, somewhat contradicting~\citeauthor{maynez-etal-2020-faithfulness}. Similarly, \citet{chen-etal-2021-factuality-checkers} analyze mispredictions on a set of previously annotated summarization corpora~\citep{kryscinski-etal-2020-evaluating,wang-etal-2020-asking,falke-etal-2019-ranking,maynez-etal-2020-faithfulness}.
The study identifies common error types (e.g., ``Numerical inference'') and constructs an adversarial test set 
with rule-based transformations. The diversity of approaches in the literature shows that evaluating factual truth is (perhaps unsurprisingly) a complex, ill-defined, and unsolved task. 
Additionally complicating this problem is that artificially introduced errors rarely match errors of real summarization models, which means that metrics trained on synthetic errors may not generalize to real systems~\citep{goyal-durrett-2021-annotating}.


Researchers have studied the validity of faithfulness metrics for other NLG tasks as well. For table-to-text, \citet{thomson-reiter-2020-gold} report the performance of an information extraction-based metric~\citep{wiseman-etal-2017-challenges} given different types of errors, and highlights typically problematic cases such as errors with names and numbers which are not detected by the metric.
Taking all these points into consideration, we conclude that there is no consensus on how best decompose and measure faithfulness and that even the best current approaches are typically flawed. However, we can also see a clear benefit to measuring specific aspects of output quality and thus encourage metric designers to stop treating output quality and in particular faithfulness like a one-dimensional problem.

\paragraph{Parameter Choices and Reproducibility}

Despite these findings, most publications still use only a single metric to demonstrate improvements over prior systems. For example, 100\% of papers introducing new summarization models at *CL conferences in 2021 use \textsc{ROUGE} and 69\% use only \textsc{ROUGE}. It thus warrants a deeper look into \textit{how} \textsc{ROUGE} and other metrics are used.

The most commonly reported \textsc{ROUGE} configurations are the F1 scores of \textsc{ROUGE}-1, -2, and -L. This choice was initially popularized by \citet{rush-etal-2015-neural}, who picked a subset of the options used in DUC 2004 which also included 3, 4, and LW~\citep{over2004introduction}. However, this choice was not empirically motivated, and from DUC 2005 onwards, the recall scores of \textsc{ROUGE}-2 and \textsc{ROUGE}-SU4 were even used instead~\citep{dang-2006-duc}.\footnote{Note though that DUC 2005 evaluated query-focused summarization instead of sentence compression which was the task studied by \citet{rush-etal-2015-neural}.} On top of the disconnect between the past and present choices, both of them are actually suboptimal. 
\citet{rankel-etal-2013-decade} find that rarely used configurations of \textsc{ROUGE} are outperforming commonly used one, and in an investigation of all 192 \textsc{ROUGE} configurations, \citet{graham-2015-evaluating} find that none of them outperformed \textsc{BLEU} and that best performance was achieved with the precision variant of \textsc{ROUGE-2}.
The studies by \citet{kryscinski-etal-2019-neural} and \citet{Fabbri21summeval} evaluate the F1-variants of multiple \textsc{ROUGE} versions and confirm the suboptimal setting. They find that \textsc{ROUGE}-1, -2, and -L perform strictly worse than \textsc{ROUGE}-3, -4, and -WE-1 across multiple rating dimensions.

Beyond using a suboptimal setup, additional parameters are often unclear; the most popular Python implementation, for example, uses a different list of stopwords compared to the original PERL script,\footnote{The package can be found \href{https://github.com/google-research/google-research/tree/master/rouge}{here}. Anecdotally, wrappers around the original implementation can lead to changes of more than 0.5 points.} but implementation details are rarely specified. That means that not only do we rely on a metric that consistently underperforms others, we are not even using it correctly or in a replicable manner. Beyond versioning issues, \textsc{ROUGE} was initially designed to evaluate English text, and it thus uses whitespace tokenization, and and English stemmer and stoplist. Yet, it is commonly applied to other languages without mentions of the exact changes to get it to run.

Similar issues exist in modern frameworks as well, especially those that utilize pretrained models~\citep{liao2021are}. For example, \textsc{BERT-Score}~\citep{zhang20bertscore} is reported in many recent summarization publications, but the term \textsc{BERT-Score} refers to the methodology instead of underlying model. To combat the confusion between model versions, the library produces a unique hash, inspired by the \textsc{SacreBLEU} framework~\citep{post-2018-call}. Yet, these hashes are often not reported or aggregated in incomparable ways.\footnote{For example, \href{https://paperswithcode.com/sota/machine-translation-on-wmt2014-english-german?metric=SacreBLEU}{Papers With Code for WMT 2014 en-de} compares models on \textsc{SacreBLEU} score without hashes.}

Another example of an often unreported design choice is how to use single-reference metrics in multi-reference setups. While \textsc{ROUGE} explicitly describes how to use it in multi-reference tasks,\footnote{The multi-reference version of \textsc{ROUGE} represents a very generous upper bound in which results can only improve by adding a reference, never decrease, which can have other negative implications. Moreover, not all implementations may use the originally recommended method.} most neural metrics do not. For example, \textsc{BLEURT}~\citep{sellam-etal-2020-bleurt} only suggests taking the max of multiple scores without discussing tradeoffs compared to computing the mean.\footnote{The alternative approach can be seen on the leaderboard of the ToTTo dataset~\citep{parikh-etal-2020-totto} where the mean of multiple \textsc{BLEURT} scores is reported.} All these evaluation parameters can have a drastic influence over the validity of scores and can lead to incorrect comparisons or inflated scores.

\subsection{Do Benchmarks Help?}
\label{sec:automatic-mt}

To develop reliable metrics, it may be helpful to develop benchmarks to collect large-scale annotated evaluation data, which may then be used to train better metrics. This has been the approach in MT for over 15 years~\citep{koehn-monz-2006-manual}, with metrics shared tasks organized as part of the yearly WMT workshop/conference. They have led to improved human annotation processes and metrics evaluation approaches, as well as almost all the learned metrics listed in section~\ref{sec:automatic-metrics}. 
As part of these shared tasks, \citet{machacek-bojar-2014-results} and \citet{stanojevic-etal-2015-results} used non-expert crowdworkers to perform a 5-way comparisons between systems. However, they point out that 5-way comparisons are challenging to interpret as \textit{pairwise} comparisons, which is required to compute \textit{segment-level Kendall-Tau correlations}.

Addressing this issue, \citet{bojar-etal-2016-results} experimented with three measuring techniques: the original 5-way ranking, \textbf{direct assessments} (DA) where outputs are evaluated by themselves, and \textsc{HUME}, a method which aggregates scores for semantic units.
After promising results, \citet{bojar-etal-2017-results} only used DA on a 0-100 scale and \textsc{HUME}. To compute correlations, DA annotations were converted into relative rankings, called \textsc{DARR}. The following year also abandoned \textsc{HUME} and fully relied on DA~\citep{ma-etal-2018-results}, and embedding-based metrics started strongly outperforming other metrics.
The 2019 shared task introduced a quality estimation task in accordance with the DA data collection technique, illustrating how the human evaluation techniques can influence the design of metrics~\citep{ma-etal-2019-results}.

However, as metrics and systems improved further, the DA annotations proved insufficient to identify a ``best'' metric~\citep{mathur-etal-2020-results}, which led to another major change to the methodology~\citep{freitag-etal-2021-results}. The latest evaluations thus followed the suggestion by \citet{freitag21experts} to use Multidimensional Quality Metrics ~\citep[MQM,][]{lommel2014multidimensional}, a fine-grained expert-based annotation approach. The results demonstrate that DA is unreliable for high-quality translations, often mistakenly ranking human translations lower than system outputs whereas human translations are correctly identified as better than system outputs in MQM. Surprisingly, metrics correlate much better with MQM, even those trained on the DA annotations.

Does this mean that focusing on DA was wrong? No, without many years of (suboptimal) data collection, we would not have learned metrics, and we would not know whether DA worked for MT. However, the progression also teaches the lesson that benchmarks may lead the field down the wrong path. A similar argument by \citet{DBLP:journals/csl/Hirschman98} critiques that benchmark evaluations only take a narrow approach and states that evaluation is intrinsically a cost-benefit trade-off. They further argue that we should weigh the divergent needs of stakeholders when designing evaluations, similar to \citet{ethayarajh-jurafsky-2020-utility}, who argue that not everyone may derive the same utility from an improvement on a leaderboard. \citet{scott2007nlg} warn that NLG evaluation shared tasks could harm the field, since they may amplify issues with the data and that benchmarks may lead to people to ignore external evaluations, and put too much emphasis on metrics that do not measure what we think they measure, both of which also happened. We thus can conclude that benchmarks are necessary, but that they need to be self-critical and explore different evaluation approaches.
\footnote{We also note that, in addition to DUC/TAC, there has been a long history of shared tasks in the NLG community addressing a much more diverse set of tasks starting with referring expression generation~\citep{gatt-etal-2008-tuna}, but which have also covered tasks such as summarization~\citep{syed-etal-2019-towards} and data-to-text generation~\citep{Dusek20evaluating}.}

\subsection{Auditing and Interpreting Metrics}
\label{sec:automatic-audits}

As seen through the WMT metrics shared tasks, machine learning-based metrics are promising, but a common criticism is that they are not transparent; it is often unclear how they operate internally and whether they can deliver high performance consistently across domains, tasks, and systems. Metric developers typically report agreement with human ratings on specific test subsets filtered on the property of interest, or they measure the change in a metric's value when perturbing a reference (e.g., by shuffling words). The idea to write \textit{tests for metrics}, rather than reporting corpus-wide correlations, may partly be traced back to~\citet{lin-och-2004-orange}, who pose that metrics should always rank a human-produced reference first when compared to multiple system outputs and thus measure how far the reference deviates from the first spot.\footnote{As we discuss later, this strong assumption is rarely met for NLG datasets.}

This section gives an overview of various research efforts that seek to evaluate automatic metrics experimentally, with each focusing on a specific aspect of the metric, such as its sensitivity to sequence length or to lexical overlap between the candidate and the reference.

\textbf{Perturbation Analysis and Surrogate Models} One common methodology is to apply methods from the interpretability literature to understand what metrics focus on. In one such study, \citet{Kaster21Global} measure to what extent several BERT-based metrics correlate with a simple linear model based on hand-crafted features. They find that these metrics are sensitive to lexical overlap despite the fact that the initial motivation for distributional similarity metrics was the over-reliance on lexical overlap of \textsc{BLEU} and \textsc{ROUGE}. The authors craft adversarial examples, and show that metrics can be fooled by lexically similar, non-paraphrase sentences. To the same end, \citet{Sai21Perturbation} conduct a correlation analysis after applying 34 perturbations that test the metrics' sensitivity to task-specific criteria (e.g., jumbling word order, introducing spelling errors for \emph{fluency}, or changing numbers for \emph{correctness}) using the Checklist method~\citep{ribeiro-etal-2020-beyond}. The results of this analysis, which covers 18 criteria across six tasks, indicate that trained metrics tend to do better, but tuning towards overall quality across task is a poor practice, leading to metrics that evaluate no individual aspect correctly. \citeauthor{Sai21Perturbation} further report that even metrics that score highly are not entirely robust to simple perturbations, calling for a more widespread use of this type of analysis.

Aside from lexical overlap, another aspect of text that has been shown to confound metrics is length. During the DUC summarization tasks, systems were restricted to a strict number of output bytes and thus were compared at a given length. This is no longer the case in modern datasets, but \citet{sun-etal-2019-compare} show that this can have dire consequences. Specifically, up to a certain length, one can ``cheat'' \textsc{ROUGE} scores by simply generating longer outputs. Even when the longer outputs are qualitatively worse, scores increase.



\paragraph{Impact of the Systems' Quality} As models improve, so should metrics. Yet, many metrics are tuned or benchmarked using previously published system outputs, which cannot be representative of the current and future state-of-the-art. As a result of this, \citet{peyrard-2019-studying} find that summarization metrics with previously reported high correlations with humans disagree with one another when tasked to compare high quality summaries, revealing their fragility. \citet{bhandari-etal-2020-metrics} revisits this conclusion, demonstrating that metrics disagree whenever the quality range is narrow, regardless of whether the summaries are good or bad. \citet{bhandari-etal-2020-evaluating} also highlight that previously published studies of metrics would yield different conclusions with more recent datasets and top scoring systems, and that the relative performance of metrics vary a lot across datasets.
These studies show that it is still unclear how metrics generalize across time, systems, and datasets and the evaluation of such qualities is complicated due to the cost of collecting human annotations, the low diversity of existing datasets, and the impossibility to to access future systems. 

\subsection{Takeaways for Metric Developers}
Since \textsc{BLEU} was introduced, dozens of papers have shown that automatic metrics have poor correlations with human judgments of quality (in addition to those cited above, see, e.g., \citet{callison-burch-etal-2006-evaluating}). We challenge the premise that such a correlation would be desirable, because quality is a vastly under-defined property. Instead, we make the case for multi-dimensional evaluation.
This is already common in human evaluations; researchers often collect evaluations for several aspects of a generated text's quality (e.g., in MT, rating both the fluency and adequacy of a translated text).
Since a single number cannot give an accurate depiction of system's performance, we call for the development of metrics with a smaller, but better defined scopes.

Another aspect that does require more attention is robustness. Meta-evaluation studies have shown that metrics can behave vastly differently on different datasets and when tasked to evaluate different NLG systems. Furthermore, multiple studies demonstrate that automatic metrics easily break when the input is subject to simple perturbations. This shows that there is major headroom for improvement: the metrics should be narrower in the phenomenon they try to capture, but broader in the input domain on which they perform well.

Given the results reported on existing benchmarks, we support the view that human evaluation remains an essential component of performance analysis, complementary to automatic metrics. 
In addition, collected annotations, especially non-English ones, may be used to train future metrics, feeding the positive feedback loop that ties metrics, models, and human evaluation.

\section{Challenges of Human Evaluation}
\label{sec:human}

The work presented in the previous section concludes human evaluation is a necessary component of model evaluations since we cannot trust automatic metrics. This conclusion is reached by treating human evaluation annotations as the ground truth to which automatic metrics are compared, and human annotations are also used as training corpora for automatic metrics. We thus rely on human evaluations and often treat them as a panacea that reveals the ultimate truth about NLG system performance. Yet there are deep-running issues with how human evaluations are conducted, which affect these system analyses, metric evaluations, and newly developed metrics. 

\subsection{What is Measured?}

While some work asks evaluators to rate the overall quality of generated text, it is more common to collect evaluations for specific dimensions of text quality. However, there is little consensus on which dimensions to evaluate. 

In the human evaluations analyzed in \citet{howcroft-etal-2020-twenty}'s study of 165 NLG papers, generated text was evaluated along 204 dimensions of quality, which they mapped to 71 distinct criteria. Some of these criteria are hierarchical, e.g., \textit{grammaticality} and \textit{spelling} fall under the more general \textit{correctness of surface form} criterion.
There are also cases where researchers apply the same text quality dimension differently. For example, \citet{howcroft-etal-2020-twenty} found that what researchers called \textit{fluency} could actually be divided into 15 different criteria, depending on how the term was defined and used in the context of the task. 

The disparities in how text quality dimensions are applied and defined in human evaluations complicate comparisons across efforts and benchmarking improvements over previous work.
This problem is exacerbated by the lack of human evaluation details in NLG papers. Of the 478 quality evaluation questions studied by \citet{howcroft-etal-2020-twenty}, over 50\% did not define the criterion they were evaluating for (279 out of 478), 65\% did not report the exact question they gave the evaluators (311/478), and 20\% did not even name the criterion being evaluated (98/478).
To promote more standardized human evaluations, some researchers have proposed detailed definitions and methodologies for human evaluation for a specific task and/or dimension of text quality. For example, \citet{thomson-reiter-2020-gold} propose a methodology for evaluating accuracy for data-to-text generation tasks, and \citet{DBLP:journals/corr/abs-2112-12870} define a framework for evaluating whether generated text is attributable to identified sources.

While general or vague evaluation criteria can lower the reproducibility and lead to low agreement between evaluators, well-specified human evaluation comes at a cost. For example, the human evaluation protocol used in the accuracy shared task at INLG 2021 \citep{reiter-thomson-2020-shared, thomson-reiter-2020-gold} produced high inter-annotator agreement, but \citet{DBLP:conf/inlg/ThomsonR21} reported that each 300-word text took an annotator 20-30 minutes to evaluate and the annotation cost for a single generated text was about US\$30. However, this detailed human evaluation protocol captured error categories that the automatic metrics were unable to detect.

\subsection{How is it Measured?}
Previous work indicates that the way questions are framed, the types of text that are being evaluated, and the measurement instruments can affect the results of human evaluations.
\citet{schoch-etal-2020-problem} discuss the role cognitive biases can play in the way researchers elicit human evaluations, such as using positive or negative framing (e.g., \textit{How much more fluent is sentence A vs. sentence B?}), including text artifacts or study design details that reveal the researchers' hypothesis, and framing instructions and questions around a model's known strengths and weaknesses. \citet{choi2005peer} provide a longer catalogue covering 48 of these biases. However, if researchers do not report the details of their studies, no one can judge whether any of these biases would apply; surveys of NLG papers find as few as 35\% \citep{howcroft-etal-2020-twenty} and 16\% \citep{schoch-etal-2020-problem} of papers share the questions used in their human evaluations.

Aspects of the texts themselves may also unduly affect the evaluators' judgments. For example, \citet{sun-etal-2019-compare} find that several dimensions of summary quality (e.g., informativeness) are correlated with the summary's length and thus suggest normalizing for summary length when evaluating these criteria. \citet{bhandari-etal-2020-evaluating} find that the relative quality of the generation models also makes a difference, showing significant differences between older annotations and newly collected human judgments for better models.\footnote{However, this finding may be confounded by the collection approach as well~\citep{shapira-etal-2019-crowdsourcing}.} They show that automatic metrics trained on annotations of text generated from older models do not always perform as well when evaluating state-of-the-art generated text. Another confounder, which we point out in section~\ref{sec:automatic}, is the correlation between dimensions that should not be correlated. \citet{Dusek20evaluating} demonstrate that the correlation can be avoided by running different annotation tasks in parallel, but this leads to a much higher cost to the evaluators. 

\paragraph{Measurement instruments}
\citet{DBLP:journals/csl/LeeGMK21} find that Likert scales were the most popular method for rating generated text, used in 56\% of studies (82/147). 
However, \citet{belz-kow-2010-comparing} argue that rating scales like those used in direct assessments (i.e., evaluating a generated text alone, without referencing other candidates) have many issues: they are unintuitive, agreement numbers are low, and most statistical measures are inappropriate for ordinal data. They find that these issues can be addressed to some extent by switching to preferential judgments.
\citet{kiritchenko-mohammad-2017-best} demonstrated that best-worst scaling (asking evaluators to choose the best and the worst items in a set) is an efficient and reliable method for collecting annotations, and this approach has been used to collect comparative evaluations of generated text \citep[e.g.,][]{liu-lapata-2019-hierarchical, amplayo-etal-2021-aspect}.

\citet{belz-kow-2011-discrete} further compare continuous and discrete rating scales and found that both lead to similar results, but raters preferred continuous scales, consistent with prior findings~\citep{svensson2000comparison}.\footnote{One potential caveat is that these studies were conducted before the wide availability of crowdsourcing platforms and are thus conducted with small cohorts of raters who have a different motivation.} 
Contrary to these findings, \citet{bojar-etal-2016-results} and \citet{novikova-etal-2018-rankme} compare direct assessments and relative rankings and find that the rankings produced were very similar, but \citeauthor{novikova-etal-2018-rankme} conclude that relative rankings are best when combined with magnitude estimates.
They also find that collecting judgments in \textbf{separate tasks} decorrelates different evaluation criteria, albeit at a higher cost since multiple tasks have to be run.

\subsection{Statistical Significance}
\label{sec:human-statistical}
Human evaluations present yet another issue: how to measure the significance of human evaluation results? \citet{DBLP:journals/csl/LeeGMK21}'s survey finds that only 23\% of NLG papers report statistical analyses to determine the significance of their results, and only 13\% explicitly state their hypotheses.

One challenge when testing for significance in human evaluation results is small sample sizes; given that the median number of generated texts in a human evaluation is 100 items \citep{DBLP:journals/csl/LeeGMK21}, most typical experimental designs for human rating studies will be underpowered to detect small model differences. This problem is not specific to NLG. \citet{card-etal-2020-little} analyze popular NLP datasets and find that they are not adequately powered (e.g., a typical MT test set of 2000 sentences would have approximately 75\% power to detect differences of 1 BLEU point). \citet{howcroft-rieser-2021-happens} demonstrate that treating ordinal data as interval data makes tests even more underpowered, which is what most papers do when analyzing rating and Likert scales (68 out of 85 recent papers, according to \citet{amidei-etal-2019-use}). Significance thresholds are not always adjusted when running multiple significance tests (e.g., Bonferroni correction), increasing the likelihood of false positives \citep{van-der-lee-etal-2019-best}.

Improvements in NLG models also make detecting statistically significant differences more challenging. Text generated by high quality models may differ less often or in more subtle ways, which requires more human judgments to detect. \citet{wei-jia-2021-statistical} show that the requirement for more judgments can quickly becomes prohibitive: to detect a difference of 1 point on a 1-100 scale in WMT, we need 10,000 perfect annotator judgments. As a result, they suggest that automatic metrics may actually be more reliable than human annotations if the annotations are insufficiently powered. The number of required annotations can potentially be decreased by not uniformly sampling examples to annotate and instead biasing the sampling toward those where models differ. However, this process can lead to artificially high correlation of the results with automatic metrics, which could overstate their effectiveness and the quality of human annotations~\citep{10.1162/tacl_a_00417}. 
Moreover, since NLG models may only differ in very few examples, statistical analyses should also handle ties as discussed by \citet{dras-2015-squibs} for pairwise rankings. 

Aside from the parameters of the study, there are also confounding factors in the evaluation of the annotation quality itself. 
To demonstrate that the annotations are of sufficient quality, reporting inter-annotator agreement is the most common method. However, \citet{amidei-etal-2019-agreement} survey 10 years of annotation agreement measures and show that almost all studies fail reliability tests. They argue that a substantial amount of the variability cannot and should not be eliminated since evaluation of generated text is intrinsically subjective and relies on many different factors including rater experience, motivation, knowledge, or education. As a remedy, they suggest using additional correlation measures alongside kappa statistics.

\subsection{Who is Measuring?}
In many human evaluations, a small number of evaluators judge the generated text. 39\% of papers in \citet{DBLP:journals/csl/LeeGMK21}'s survey use between 1--5 evaluators. However, it is becoming increasingly common to collect judgments from a large number of evaluators using crowdsourcing platforms like Amazon Mechanical Turk (MTurk), Appen, Prolific Academic, and Upwork.

In particular, MTurk has a long history in NLP with early claims stating that a small number of crowdworkers can replace a single expert rater \citep{snow-etal-2008-cheap}. Similar claims were made in other communities, stating that, while not as high-quality, overall data quality can actually be improved by having more redundant annotations~\citep{DBLP:conf/kdd/ShengPI08}. However, later studies find that this point is actually a lot more nuanced.  
Some dimensions of text quality may be easier than others to rate with crowdsourced evaluators instead of experts.
\citet{gillick-liu-2010-non} find that MTurk judges were better at measuring generated summaries' linguistic quality than their content or overall quality and had a much higher correlation between linguistic and overall quality than experts.
\citet{clark-etal-2021-thats} find MTurk evaluators are more likely to base judgments of generated text on the text's form rather than its content.
In their work on German summarization evaluation, \citet{iskender-etal-2020-best} find that non-redundancy and usefulness are very hard to assess using crowdworkers and suggest that experts should be used for them, while crowdworkers are suitable for other dimensions of text quality as long as results are carefully interpreted. 

Analyzing DUC annotations between 2001 and 2004, \citet{harman-over-2004-effects} find that averaged human ratings can yield meaningful insights, but also note that there is very high variance both within and between human raters and that it is unclear whether the source of the variance is intrinsic to the humans or the models.
This variance may be even higher in crowdsourcing scenarios compared to expert raters. 
\citet{karpinska-etal-2021-perils} report that running the same MTurk evaluation on different days of the week can vary enough to produce different results. When analyzing evaluations of MT systems, \citet{freitag21experts} find that agreement between ratings produced by linguists and those from crowdworkers can be extremely low. In fact, they find that \textbf{automatic metrics can have higher agreement with high-quality annotations than human crowdworkers}. Some tasks like multi-document summarization are especially challenging and time-consuming for people to evaluate. Observations like these have led to work proposing evaluation methods that combine the advantages of human and automatic evaluation \citep[e.g.,][]{hashimoto-etal-2019-unifying, DBLP:conf/emnlp/ZhangB21}.

The increasing quality of generated text has led some researchers to move away from crowdsourcing platforms.
For example, expert evaluators like English teachers \citep{karpinska-etal-2021-perils} or trained, in-person evaluators \citep{ippolito-etal-2020-automatic} were needed to distinguish between human-authored text and text generated by today's generation models (an evaluation most commonly found in dialogue generation).
Similarly, \citet{freitag21experts} demonstrate that \textbf{non-expert annotations often lead to mistaken claims of super-human model performance}, when expert annotators correctly identify issues in the generated texts. 

It is unclear whether these issues are specific to the fact that non-expert annotators are being used, or if these issues may be overcome by improving the quality of the study and the working condition of raters. 
Investigating the use of MTurk for NLP, \citet{DBLP:journals/corr/abs-2111-05241} find that about 25\% of studies have technical issues, 28\% have flawed, vague, or insufficient instructions, and 26\% of study creators were rated as having poor communication. Notably, they also find that 35\% of requesters pay poorly or very badly according to MTurk raters. 
To that end, many have questioned whether the treatment evaluators receive and the structure of crowdsourcing platforms provide ethical working conditions for evaluators. The most basic of these considerations is payment; does the low-pay, small-batch format of crowdsourcing actually provide evaluators with a fair wage? \citet{10.1162/COLI_a_00057} discuss the low wages MTurk workers receive, along with concerns about data quality issues that the platform incentivizes. These concerns are not unique to MTurk; \citet{DBLP:conf/cgc/Schmidt13} argues that there are ethical concerns across crowdsourcing platforms, regardless of how they incentivize workers. \citet{shmueli-etal-2021-beyond} cover a broader set of ethical considerations for crowdsourcing work, including potential psychological harms, exposing sensitive information about workers, and breaching workers' anonymity. Despite these concerns, \citeauthor{shmueli-etal-2021-beyond} report that only 14 out of 703 NLP papers that used crowdsourcing mention IRB review.

\subsection{Subjectivity and User Satisfaction}
Most of the human evaluations in this section are intrinsic evaluations, asking evaluators to rate the quality of the generated text. However, the more valuable question is answered with extrinsic evaluation: \textbf{how well does the generated text serve its intended purpose?} These evaluations measure how useful a text generation model is and indicate whether real world users would be satisfied with the generated texts. Evaluations focused on intrinsic qualities of the text fail to capture dimensions of NLG systems that practitioners care about, e.g., how trustworthy a generated text is or how well it performs in human-in-the-loop settings.\footnote{See, for example, \href{https://ehudreiter.com/2021/09/27/inlg-what-real-world-users-want/}{Ehud Reiter's summary}  of a panel on NLG in industry at INLG 2021.}

Another related aspect that is rarely considered in human evaluations is the subjectivity of text evaluation. People may value certain text qualities more highly than others or be working from a different point of reference. Even the more ``objective'' aspects of text quality, like grammatical correctness, may depend on the evaluators' dialect, the perceived formality of the text, the context or style of the generated text, etc. Disagreement in evaluators' ratings does not always indicate evaluator error; rather it may be a signal that there is more complexity to the text or dimension of quality.
While it has been shown that increasing the number of annotations per example can decrease the overall bias~\citep{artstein2009bias}, this finding assumes that the population of annotators is somehow representative of the whole world. 
\citet{prabhakaran-etal-2021-releasing} find that \textbf{aggregating annotator responses results in under-representation of groups of annotators' opinions}, and they recommend releasing annotator-level annotations and collecting annotators' socio-demographic information to prevent the exclusion of minority perspectives.
We thus should be careful of results such as those that suggest excluding data with low agreement scores with other annotators~\citep{owczarzak-etal-2012-assessing}, unless we know the source of the disagreement is not subjectivity. 
Even well-established NLG tasks have aspects of subjectivity that are usually ignored. For example, the goal of a summarization task is to generate the important points from a document, but \citet{kryscinski-etal-2019-neural} find that when annotators select which sentences in a document are the most important to include in a summary, the majority of evaluators only agree on an average of 0.6 sentences per document.

While the majority of evaluation criteria is by definition subjective, there is an opportunity for hybrid approaches with the help of standardized measures~\citep{DBLP:journals/csl/LeeGMK21}. One such dimension that could be useful for tasks like simplification is the readability of text, which could be measured using scales such as the ones proposed by \citet{kincaid1975derivation} or 
\citet{ambati-etal-2016-assessing}. \citeauthor{DBLP:journals/csl/LeeGMK21} point out that the relationship between these objective measures and subjective readability assessments is not currently being studied, although a strong objective measure could lead to a higher degree of standardization. Similarly, one can imagine human-in-the-loop approaches for measuring faithfulness that focus on claims that are challenging to verify using only automatic approaches, enabling the collection of a much larger quantity of judgments.

\section{Challenges with Datasets}
\label{sec:data}

A component mostly kept apart from evaluation analyses is the data, even though NLG tasks are embodied through datasets; for example, claims about performance on CNN/DM may be used as a proxy for performance on all summarization tasks. Issues with datasets are widely studied in the general machine learning literature which we heavily draw on in this section, with anecdotal evidence for NLG tasks when available. In a recent survey of datasets and benchmarks in machine learning, \citet{liao2021are} point out that the lack of differentiation between tasks and datasets that aim to capture them can lead to harmful over-generalization. They argue that choosing to evaluate on a dataset reinforces design decisions taken during its construction and focuses the evaluation on the specific distributions represented in the data.

Collectively, the research community could select for a more diverse language representation and decide to replace older flawed datasets by newly developed ones. 
Unfortunately, the collective choices also reinforce suboptimal design decisions. Analyzing a sample of 20 papers that proposed summarization approaches in 2021, we find 27 datasets that models were being evaluated on. The most popular ones, CNN/DM and XSum~\citep{narayan-etal-2018-dont}, were used five and four times respectively, despite their issues, which we explore in section~\ref{sec:data-examples}. Additionally, \textbf{only two of the 27 datasets were non-English}, despite much recent work that introduces multilingual summarization corpora~\citep{giannakopoulos-etal-2015-multiling,scialom-etal-2020-mlsum,ladhak-etal-2020-wikilingua,hasan-etal-2021-xl,perez-beltrachini-lapata-2021-models}.

These findings lead to three questions. First, how can we as a research field measure summarization improvements on disjoint datasets? How can we claim that we are making progress if we only focus on a single language? And, given the significant issues with popular benchmark datasets, what do improvements even mean? Throughout this section, we analyze typical design choices during NLG data construction and how they influence insights derived from evaluations.\footnote{We point to \citet{DBLP:journals/corr/abs-2012-05345} for a more in-depth survey of general issues in data creation, including those of benchmarking and data maintenance practices, to \citet{DBLP:conf/fat/BenderGMS21} for a survey issues of using large web-scraped datasets, and to \citet{luccioni-viviano-2021-whats} and \citet{dodge-etal-2021-documenting} for analyses of such large-scale web-scraped corpora and their representational, legal, consent, and PII issues.}

\subsection{Representation in Performance Numbers}
\label{sec:data-representation}

Dataset creation is a value-laden process, yet those values are rarely made explicit~\citep{DBLP:conf/fat/HutchinsonSHDGK21}. The choices of dataset creators have significant impact, for example on who is represented in the data and on the language(s) of a dataset. \citet{joshi-etal-2020-state} assess the language diversity in NLP, showing that very few languages beyond English are being studied, regardless of the number of their speakers. A similar argument can be made for dialects; focusing on African American Vernacular English (AAVE), \citet{blodgett-etal-2020-language} describe multiple studies showing a drop in performance on popular NLU tasks when applied to text with AAVE features~\citep[][among others]{jorgensen-etal-2015-challenges,jorgensen-etal-2016-learning,blodgett-etal-2016-demographic}. Beyond performance drops, excluding dialects from datasets can often be seen as akin to de-legitimizing the language and their speakers~\citep{Rosa2017UnsettlingRA}.
This problem is even worse in NLG, where no popular corpora exist to measure the discrepancy in performance between dialects, and, as seen above, the most popular corpora only cover versions of English present on popular British or US news websites. When making claims about model performance, we should thus acknowledge that we report it for only a tiny sliver of possible phenomena and work toward reporting performance for different subpopulations~\citep{DBLP:conf/fat/MitchellWZBVHSR19}.

\paragraph{Design Choices} Beyond actively reporting more fine-grained numbers, \citet{DBLP:conf/fat/HutchinsonSHDGK21} propose that the assumptions underlying a dataset should be specified \textit{before and during the collection} to enable an early peer review of the choices. Instead of releasing datasets as monolithic artifacts and treating them as number-producing black-boxes, they should be accompanied by sensitivity studies for dataset parameters and rigorous discussions of their limitations. Unfortunately, none of these suggestions are typically followed: \citet{DBLP:journals/pacmhci/ScheuermanHD21} analyzed 114 computer vision datasets and find that their creation process values efficiency at the expense of care and that they typically aim to be as universal as possible without nuanced understanding of contexts from which datapoints arise. All this typically benefits the model work at the expense of data work, leading to easier-to-digest but deeply flawed results, similar to what we have discussed so far for NLG evaluations. 
Similarly, through interviews with 53 AI practicioners, \citet{DBLP:conf/chi/SambasivanKHAPA21} highlight how data collection choices cascade and amplify through all parts of the development pipeline from training and evaluation to deployment. They warn of the lack of incentives to produce high-quality datasets and encourage more work on data improvement processes that should be part of the life cycle of a dataset. 
To address some of the representational issues, it seems natural that we should aim to produce ``impartial'' data, but this may also be either undesired or even impossible. \citet{rogers-2021-changing} summarize discussions around \textbf{data curation}, the act of manufacturing distributions that differ from naturally occurring ones, pointing out that dataset creators should maybe not be the ones deciding what distribution should represent the world, and that studying the world as it is with all its flaws and biases is an important aspect of NLP. There is thus not a one-size-fits-all curation solution. 

Regardless of the choices of dataset creators, it is imperative to report the curation decisions alongside limitations of datasets in structured format to allow for a better interpretation and contextualization of performance results~\citep{bender-friedman-2018-data,gebru2021datasheets,mcmillan-major-etal-2021-reusable}. To that end, interactive tools like like \href{https://knowyourdata.withgoogle.com/}{Know Your Data}, \href{https://www.ibm.com/products/dqaiapi}{Data Quality for AI}, and the \href{https://huggingface.co/spaces/huggingface/data-measurements-tool}{Data Measurements Tool} may provide valuable insights. Since the suggested documentation and analysis processes are rarely followed, we will only be able to shed some light onto issues in NLG datasets, and note that uncovering and addressing these issues should be an ongoing process.

\paragraph{Memorization} When talking about dataset issues, we also need to consider the trend of pretraining corpora that were scraped from the web. Many NLG datasets are similarly built on top of web-scrapes (e.g., news websites for summarization datasets or Wikipedia for data-to-text datasets) and often do not contain significant post-editing steps. As a result of this, pretraining examples can be found in downstream test corpora~\citep{dodge-etal-2021-documenting,DBLP:journals/corr/abs-2107-06499}. 
Since it is impossible to remove the affected data from the training corpus after the release of a model, multiple approaches have been explored mitigation techniques. For example, \href{https://github.com/google/BIG-bench}{BIG bench} introduced a hash identifier that will allow web crawlers to ignore their data, but this approach does not work for data scraped from other sites. 
Another approach investigated by \citet{yuan2021synthbio} uses large models alongside humans to create fully artificial data. However, the authors find that even careful curation aiming to diminish bias issues leads to others which may be more subtle. As it stands, the only approaches to avoid memorization are, therefore, to not rely on web-crawled data at all or to continuously build new datasets that utilize data collected after the cutoff date for large pretraining datasets, neither of which are practical for all NLG tasks. Any performance improvements on NLG tasks based on web-data should thus be analyzed carefully for the effect of memorization and test leakage.

\subsection{Communicative Goals and Noise}
\label{sec:data-examples}

Another assumption underlying corpus-based NLG is that human references represent a gold-standard. While we discussed in section~\ref{sec:metrics-likeness} that this leads to flawed metrics of ``human-likeness'', we also use human-written references as the target to optimize during training. As \citet{reiter-sripada-2002-corpora} argue, this is a fallacy since humans disagree with each other and make mistakes, which our models will learn to replicate. It is thus crucial to understand exactly what task we are actually learning from the data, whether it corresponds to the claimed communicative goal, which potential shortcuts a model may be taking, and whether there is noise in the data that distracts from the task. 

A commonly cited shortcut in summarization is positional bias. Since most common summarization datasets are built on journalist-written news articles, they typically follow best practices to provide salient information early on in the article~\citep{grusky-etal-2018-newsroom}. Summarization datasets thus have strong positional data biases that models pick up on and which lead to inflated results if the test set has the same biases~\citep{gehrmann-etal-2018-bottom}.
If the claims made about such a model are specific to news summarization and resulting models were only used to summarize news articles written in a similar style, this may not be a significant problem. 
However, \citet{kedzie-etal-2018-content} demonstrate that controlling for positional bias drastically decreases model performances to the point where deep learning based models barely outperform much simpler approaches. Therefore, it may be helpful to also evaluate summarization models on a variety of tasks including in the non-news domain to prevent inductive model biases from inflating the results. 

Another side effect of the positional bias is that simply picking the first three sentences of a professionally written article is a strong baseline, as shown by \citet{DBLP:conf/aaai/Nenkova05} who analyze DUC-2001 data and note that \say{\textit{only one system significantly outperforms the baseline of selecting first sentences from the input articles}}. The same baseline was introduced for neural models on CNN/DM by \citet{DBLP:conf/aaai/NallapatiZZ17} who similarly find that it is extremely effective. 
This effect is especially pronounced in CNN/DM where raters even prefer the first three sentences to the summary provided in the dataset~\citep{DBLP:journals/corr/abs-2009-01325}. This is due to the fact that the design choice for CNN/DM was to pair an article with the bullet points written for it on the homepage of the respective news outlet, which worked well for its intended use as a reading comprehension dataset~\citep{DBLP:conf/nips/HermannKGEKSB15}, but does not work for summarization. This re-use of datasets for incompatible tasks, along with the concentration on very few datasets, is a worrying trend that was quantified across multiple other tasks by \citet{DBLP:journals/corr/abs-2112-01716}.

Along similar lines, datasets constructed through web scrapes may additionally contain extraneous information, such as hyperlinks or image captions. Here again, CNN/DM is a culprit~\citep{Fabbri21summeval}; since the \textbf{references were never meant to be a real summary}, there is no requirement that a reference is faithful to the source article. 
An analysis of XSum finds that over 70\% of references contain external hallucinations~\citep{maynez-etal-2020-faithfulness}. 
This finding provides an opportunity for dataset developers to improve dataset construction processes---for example, XL-Sum, a recent multilingual news summarization dataset, evaluates the faithfulness of references across 10 languages and find that in their dataset, only 25-40\% of summaries contain unsupported information, a significant decrease compared to XSum. An alternative path toward this goal is to improve datasets over time once these issues are uncovered. For example, \citet{gehrmann-etal-2021-gem} release an improved version of XSum that filters the dataset based on a faithfulness classifier trained on the data by \citet{maynez-etal-2020-faithfulness}.

Similar noise can also be found in common datasets for other NLG tasks. In WikiBio~\citep{lebret-etal-2016-neural}, which has the communicative goal to provide a short biography grounded in key-value attributes about a person, less than half of the attributes are actually realized in the reference and over half of references score very low or low in faithfulness on a 5-point scale~\citep{yuan2021synthbio}. It is surprising to see how far behind the rest of NLG is behind MT in this regard, where filtering and cleaning of scraped data is common practice and shared tasks are being held~\citep{koehn-etal-2019-findings}. 

Moreover, crowdsourced NLG datasets, for which one may expect a lower ratio of noise, are not without problems. \citet{dusek-etal-2019-semantic} find that cleaning the E2E NLG dataset~\citep{novikova-etal-2017-e2e}, for which the communicative goal is to describe a restaurant given a set of key-value attributes, led to a reduction in slot-error rate of up to 97\%, which means that failures may have incorrectly have been attributed to the model instead of the data. Similar reductions in errors were seen in task-oriented dialog as a result of improving the dataset~\citep{DBLP:conf/emnlp/BudzianowskiWTC18}. Despite these findings, few datasets follow construction processes with multiple post-editing steps to ensure a low ratio of noise~\citep{parikh-etal-2020-totto}.

These examples demonstrate the importance of identifying limitations of existing ``standard'' datasets and either replacing them with better constructed ones, or---if the limitations can be addressed---improving them over time. 
More attention should be paid to \textbf{construction processes} that aim to minimize noise, and faithfulness evaluations should be default for new datasets. 
While much of this work is by nature qualitative, automatic methods can be employed to characterize aspects like the abstractiveness or compression-ratio in summarization datasets~\citep{bommasani-cardie-2020-intrinsic}.

\subsection{Constructing Informative Test Sets}
\label{data:test-sets}

We next take a look at the choices behind test set construction. It is usually considered a best practice to create i.i.d. splits. That is, we assume that a subset of the dataset is representative of the full data distribution, and randomly split the data into training, validation, and test sets. However, this assumption may not hold, and, given the sampling bias pointed out in section~\ref{sec:data-representation}, lead to similar under-representation in the test data. As a way to make i.i.d. schemes more robust, \citet{gorman-bedrick-2019-need} propose using multiple random splits similar to cross-validation as they find that results on multiple NLU tasks change when the splitting process is changed. Following prior work by \citet{DBLP:journals/jmlr/Demsar06}, this enables computing statistical significance of numbers. However, besides the increased computational complexity in NLG, \citet{sogaard-etal-2021-need} point out that i.i.d. splits may not the correct way to characterize system performance, since the above assumption implicitly assumes that the data distribution matches the distribution a model would run on during deployment in a real-world scenario and thus argue for evaluating on samples that measure aspects, for example topics or content from certain years, that are not seen during training. They evaluate multiple not-random, \textbf{informed data splitting approaches} and find that the results vary significantly depending on how the test set was constructed. 
Considering data splits during dataset construction can thus lead to much more informative results. For example, the E2E NLG dataset was used for a shared task with a private test set that contained completely unseen attribute combinations, leading to a drop in performance~\citep{dusek-etal-2018-findings}. A similar approach was taken for ToTTo, a dataset to describe a set of highlighted cells in a table, which reports numbers for seen and unseen combinations of table columns~\citep{parikh-etal-2020-totto}. Here, results on unseen combinations are more than 60\% lower than on the seen combinations.

\paragraph{Transformations}

Another factor to consider for the construction of test sets is how to handle natural language variation. Dialectic or individual variations can entail different spelling, word order, grammar, or vocabulary. To that end, \citet{DBLP:conf/emnlp/MoradiS21} show that models are very brittle to character- and word-level perturbations. Even if the dataset creation did not consider informed splits, it is possible to create \textbf{evaluation suites}, a collection of test sets that \textit{together} yield informative insights. Building on the insights from challenge sets that avoid potential model shortcuts~\citep{mccoy-etal-2019-right}, \citet{ribeiro-etal-2020-beyond} argue that for NLU tasks, one can enumerate linguistic phenomena and expected outcomes. For example, a negation should flip the result of whether a fact is entailed by its premise, but replacing an entity in both should not.

While we cannot enumerate capabilities in a similar way for NLG, \citet{mille2021automatic} argue for informed transformations coupled with collection of additional data to enrich existing datasets. Informed transformations measure the causal effect of introducing language variation, for example changing the order of columns or replacing numbers with others, while additional test data can be used to evaluate without overlap with the training set, addressing the memorization issue mentioned above. \citet{dhole2021nlaugmenter} expand their framework to over 100 different transformations that include dialectal variations, OCR errors, and others that can be used to create more realistic scenarios.

\begin{figure}[t]
    \centering
    \includegraphics[width=\columnwidth]{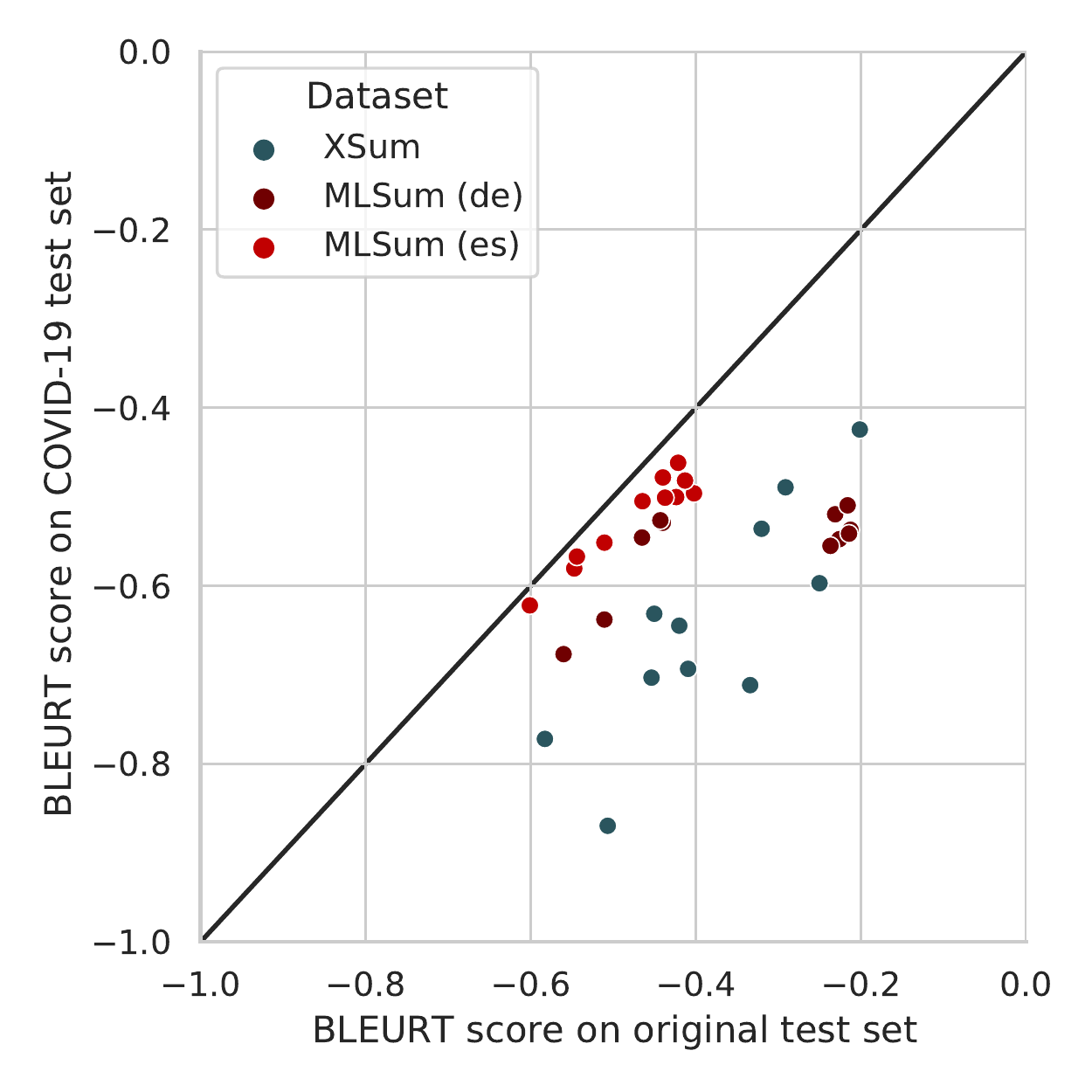}
    \caption{A comparison of \textsc{BLEURT} scores of 33 models evaluated on the original and the newly collected COVID-19 related test sets for three summarization tasks. The diagonal represents the desired model performance $f(x)=x$. The vertical distance from the diagonal indicates the extent of the performance drop when moving from one test set to the other.}
    \label{fig:covid}
    \vspace{-1.5em}
\end{figure}

\paragraph{Time Travel}
Dataset shifts are when the joint distribution of inputs and outputs differs between the development of a model and its deployment (or in our case test its test setup)~\citep{quinonero2009dataset}. One of the suggestions by \citet{sogaard-etal-2021-need} is to simulate dataset shifts to simulate a more realistic deployment scenario, which they test on the Gigaword sentence compression dataset~\citep{napoles-etal-2012-annotated} where they divide splits by year of publication. This process becomes even more important given what we know about the extent of train-test overlap in pretrained models~\citep{DBLP:journals/corr/abs-2107-06499}. 
One recently suggested approach by \citet{mille2021automatic} is to continuously collect new test sets using the original collection approach. As an example, they collect new test sets for XSum and the English and German subsets of MLSum~\citep{scialom-etal-2020-mlsum} which focus on COVID-19 related news articles which is not part of large pretraining corpora or the mentioned datasets. We re-analyze their released data for 33 models in Figure~\ref{fig:covid}, focusing on their \textsc{BLEURT} score~\citep{sellam-etal-2020-bleurt}. As can be seen, models consistently do not handle the new concept well.

\paragraph{Error Reporting} 

While evaluation suites enable quantification of errors to some extent, not all errors are detectable through general-purpose methodologies, and thus require more in-depth investigations. One such example are hallucinations, which, as shown in section~\ref{sec:automatic-audits}, are not detected by standard metrics. In addition to hallucinations, \citet{stevens-guille-etal-2020-neural} also investigate repetitions and omissions of content as plausible errors. To facilitate error analyses, \citet{higashinaka-etal-2015-fatal} propose a process of error annotations and taxonomy creation, which they demonstrate using a dialog system.
However, despite the available resources, only about 10\% of papers report any error analyses~\citep{van-miltenburg-etal-2021-underreporting}. Consequently, it is unclear what kinds and how many errors contemporary systems even make. \citeauthor{van-miltenburg-etal-2021-underreporting} extend the workflow by \citeauthor{higashinaka-etal-2015-fatal} to one which can be used to detect and quantify different error classes, and argue for stricter reporting guidelines as part of conference submissions. 
The type of errors that should be annotated should be informed by what the study sets out to explore. For example, \citet{deyoung-etal-2021-ms} investigate the problem of summarizing medical studies and their error analysis focuses on whether the effect of the study intervention was generated correctly (i.e., whether the medical study demonstrated a positive effect). This error analysis is only suited for the specific task but provides a view into the data distribution of the references and uncovers a systematic shortcoming where the system does not report the effect correctly in about 50\% of cases.

\subsection{The Nature of References}
\label{sec:data-who}

As discussed above, how a test set is constructed can have large implications on the model evaluation. Another contributing aspect is the style in which references are presented. Since we rely on human-likeness metrics in evaluation processes, the style in which the references are constructed matters significantly. An example of this is in machine translation where ``low-quality'' references that contain translationese lead to low diversity and favor translation systems that produce similar low-quality outputs~\citep{freitag-etal-2020-bleu}. This can partially be counteracted by employing experts to paraphrase existing references, thus creating a wider set of reference points to which a metric can compare, but this direction has not been explored for NLG. 

Similar problems exist in summarization, where the same references are often used to evaluate both extractive and abstractive approaches. Summaries in many datasets exhibit a high fraction of content overlap with the articles. Consequently, extractive systems are favored by design~\citep{goel-etal-2021-robustness}, and metrics have a lower correlation with human judgments as references become more abstractive~\citep{bhandari-etal-2020-metrics} due to the mismatch in style.
These findings has also been corroborated for XSum~\citep{gehrmann-etal-2021-gem,mille2021automatic}.

\section{Suggestions for NLG Researchers}
\label{sec:recommendations}

As we have seen, it is impossible to fully identify whether or how our models fail with methods available to us today. And even if we detect failure, we cannot attribute it to the data, the evaluation process, or the model itself. 
Due to the problem's complexity, it will require a significant effort to establish a positive feedback loop in which improvements to data, models, or human and automatic evaluations can benefit the other parts of this circular dependency. 
To help facilitate work toward this goal, we make the following suggestions for NLG researchers.

\subsection{Documentation, Releases and Maintenance}

Preconditions of any further progress and the better understanding of model limitations are improved documentation standards and avoidance of the documentation pitfalls discussed throughout this paper. On the data side, this includes documenting the exact data collection processes, their limitations, and a discussion of the social impact of datasets, as proposed for data cards~\citep{bender-friedman-2018-data,gebru2021datasheets}. 
Beyond identifying issues, standardized documentation following mutually agreed-upon frameworks can lower the barrier of entry to newly developed resources~\citep{lhoest-etal-2021-datasets}.
A drastic, yet necessary, change from the status quo is that datasets and their documentation must not be static entities. Datasets should be cleaned and improved~\citep[e.g.,][]{dusek-etal-2019-semantic,thomson-reiter-2020-gold} over time and sending pull requests to update data documentation needs to become as commonplace as sending pull requests to or opening issues in open-source libraries.

Additionally, treating datasets as dynamic encourages the development of evaluation suites that everyone can benefit from~\citep{bowman-dahl-2021-will}. The benefit of this approach can be seen in natural language inference, where the dataset SNLI~\citep{bowman-etal-2015-large} was extended to cover more genres~\citep{williams-etal-2018-broad} and languages~\citep{conneau-etal-2018-xnli} and subsequently has been used to explore whether adversarial data augmentation techniques are useful for evaluation~\citep{nie-etal-2020-adversarial,DBLP:journals/corr/abs-2111-08181}.
However, we also should not hesitate to take more drastic measures and deprecate datasets when better ones are released. To that end, we mirror the suggestion by \citet{bommasani-cardie-2020-intrinsic} that it is time to do that with CNN/DM, which is no longer a useful summarization dataset. 
The same goes for metrics as well: It is clear that no single metric can provide all the insights, so no paper should rely on only a single metric. Moreover, while it is too early to fully deprecate \textsc{ROUGE}, we need to normalize not reporting its scores in favor of other lexical metrics like \textsc{METEOR} that have been shown to perform at a similar or higher level. When lexical metrics are used for non-English text, the tokenization approach needs to be documented and metrics with established tokenization approaches like \textsc{BLEU} should be used in favor of \textsc{ROUGE}. For now, alongside deeper analysis, we recommend using at least one entailment or QA metric and a learned distributional similarity one like \textsc{BLEURT}, at least until we have reliable direct assessment metrics that do not require references. For Translation, we recommend to combine a lexical overlap-based metric, e.g., \textsc{ChrF} or \textsc{BLEU} (at the systems-level) with a learnt metric such as \textsc{COMET} or \textsc{BLEURT}, and encourage using of MQM for human evaluation if researchers have access to expert raters given their budgetary constraints~\cite{freitag21experts}.

\begin{figure*}[ht]
    \centering
    \includegraphics[trim=6cm 0cm 10cm 0cm, width=\textwidth]{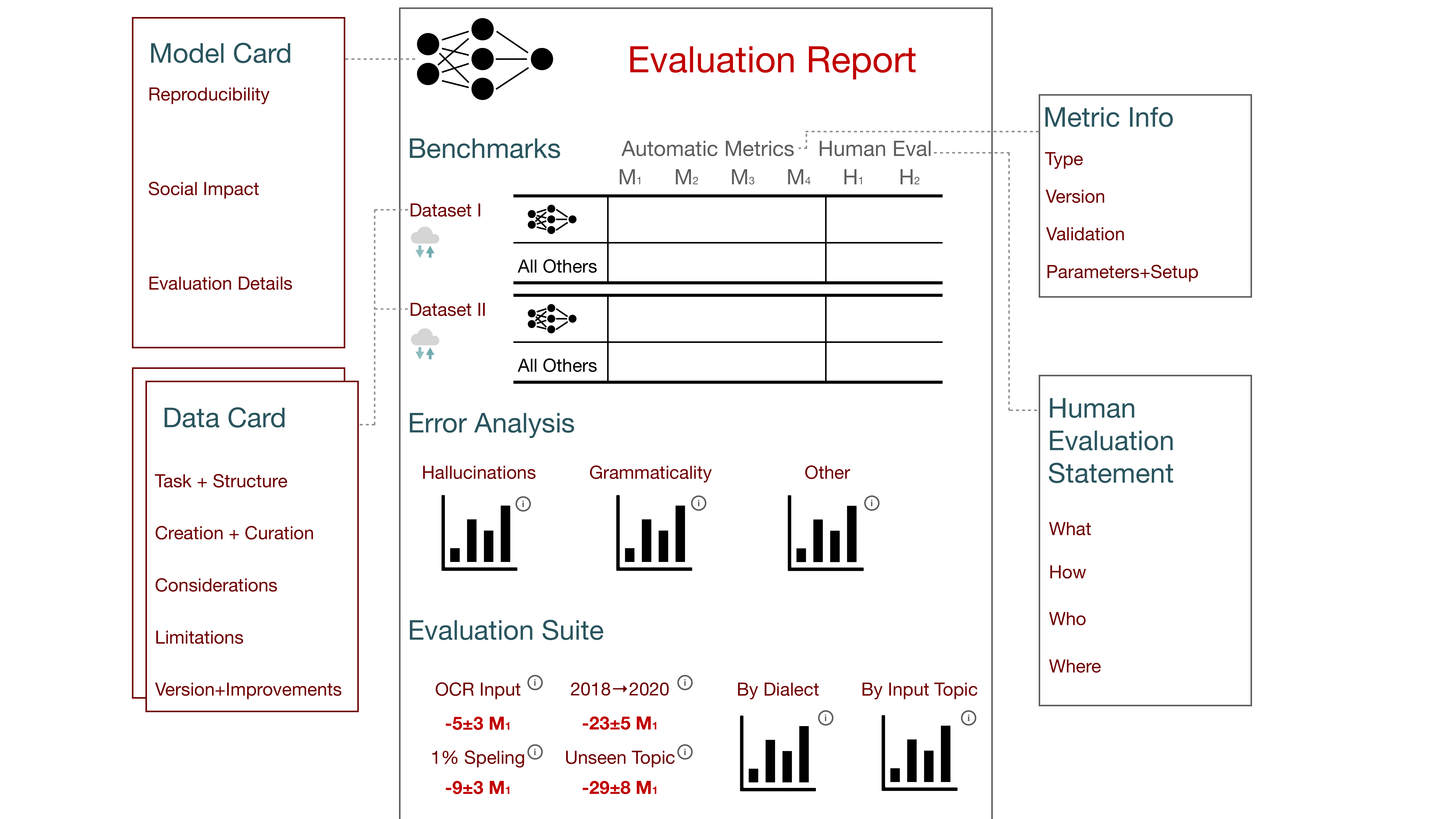}
    \caption{Our vision of an evaluation report. A model should be evaluated on multiple datasets via multiple metrics and well-documented human evaluation. The outputs, scores, and any human assessments should be easy to download. In addition, each dataset should be the most current version and accompanied by an up-to-date data card. The evaluation report next reports fine-grained error types such as hallucinations (intrinsic, extrinsic, factual), or grammatical errors (subject-verb agreement, spelling, etc.). Finally, evaluation suites are used to produce model audit results on specific input types. In this example, the model handles OCR and spelling mistakes relatively well, but fails on unseen topics or time shift. Model audits can also include breakdowns of performances by input types or dialects. The entire report is part of the model card. }
    \label{fig:eval-report}
\end{figure*}

Some existing projects focus on continuous improvements for some evaluation aspects, e.g., DynaBench~\citep{kiela-etal-2021-dynabench} aims to improve data alongside models, and Bidimensional Leaderboards~\citep{DBLP:journals/corr/abs-2112-04139} for improving metrics alongside models. 
However, DynaBench focuses on NLU and Bidimensional Leaderboards use CNN/DM as the only non-MT NLG dataset. 
It is thus unclear whether a single shared framework that addresses only a subset the mentioned issues is the solution instead of uniting the decentralized research community behind this shared goal. 

Implementing and popularizing these changes in the community will require several changes to peer review processes.
First, we should encourage authors to submit resource papers. As \citet{rogers-augenstein-2020-improve} point out, resource papers are already underappreciated and increasing what counts as acceptable documentation for a resource paper may lead to fewer such papers being written.
Second, authors and reviewers need to move from claiming empirical improvements toward a more rigorous documentation of how those were achieved. Modeling papers often include deliberations why certain architecture choices were made, but the choice of which datasets to evaluate on or which metrics are being used rarely move beyond ``other people use it''. By the same logic, reviewers may be hesitant to accept claims when a model is not evaluated on the standard flawed datasets. As discussed in this work, many of the standard practices should be reconsidered and we thus need more elaboration on these choices.
Third, we encourage researchers to focus on specific phenomena, rather than overall quality. Instead of treating NLG models or metrics as ``one big problem'', we encourage work on more specific aspects, say, logical consistency in dialog, or aggregations in table-to-text generation. We further encourage researchers to use task-specific metrics and be upfront with the trade-offs, and we encourage reviewers to expect and accept more nuanced claims and contributions while discouraging claims about the overall quality of a system.
Finally, to support this research, we should encourage re-training and/or re-implementing prior work for the most appropriate benchmark task(s) and evaluation process when necessary.


Beyond this, one of the best ways to contribute to improving evaluation as a modeling researcher is to release of model outputs for validation and test sets alongside instructions on how to replicate reported numbers. 
Many works like that of \citet{Fabbri21summeval} would not be possible without access to model outputs, and such corpora can be used for metric development and validation, and to conduct meta evaluations. 
Releasing outputs on non-English datasets, even when no human evaluation can be conducted, will ease the path for evaluation improvements on the covered languages by reducing the burden on the evaluation researchers to produce the outputs.

\subsection{Better Human Evaluations}

While human evaluation can solve many issues of automatic metrics, it often does not. We thus should not blindly accept results that show that one number is larger than another. Again, it is more important that the process to arrive at these numbers is well documented, that people involved in the process are considered, and that results are sufficiently statistically powered. 
To achieve this goal, we need to work toward reusability and replicability in human evaluations, for example by filling in human evaluation datasheets~\citep{DBLP:journals/corr/abs-2103-09710, belz-etal-2021-reprogen} and by contributing and using projects that standardize parts of the process~\citep[e.g.,][]{DBLP:journals/corr/abs-2101-06561,gehrmann-etal-2021-gem}.
Expanding on the suggestions by \citet{van-der-lee-etal-2019-best}, we need a wider adoption of effect size estimates, power analysis, statistical significance tests, and emphasize the importance of analysis of the validity of human evaluation results.

Additionally, even though much of recent advances in NLP have been powered by non-expert crowdraters, the importance of expert raters is becoming increasingly clear, both in dataset construction and model evaluation. There needs to be a clear understanding what the required qualifications are to participate in a task and whether the involved raters fulfill them.

We further suggest more work on reducing the barrier for the deployment of extrinsic evaluations. Extrinsic evaluations hold generated text to a higher standard, requiring a generated text to not only be ``correct,'' but also to be effective at its intended purpose and to be useful to a potential end user. Grounding the evaluation in a specific task and situational context moves the focus of the evaluation from the appearance of the generated text to the content and the purpose of the text.

\subsection{Model Audits and Evaluation Reports}

While ranking models according to a single quality number is easy and actionable---we simply pick the model at the top of the list---it is much more important to understand when and why models fail. A model being on top of a well-established benchmark only means that it performs best on the majority of test examples, but as we have seen, the construction of test sets is often not representative of performance on real scenarios and can hide issues in less frequent classes. 

\citet{DBLP:conf/fat/MitchellWZBVHSR19} describe the ``quantitative analysis'' process of testing on subpopulations and reporting disaggregated results according to chosen metrics, falling back on synthetic data when necessary. Our suggestion goes beyond this to create what we call \textbf{evaluation reports} as part of model cards which document the results of \textit{model audits}, as outlined in Figure~\ref{fig:eval-report}. 
The idea of an model audit is to identify what breaks a model, with the goal of moving away from chasing the highest overall number. 
The long-term goal of evaluation reports are performance guarantees: we would like to know exactly what to expect of a model for a given input. 
Since the space of potential model shortfalls is rather extensive, the creation of model audit processes will rely on our collective work to create evaluation suites and on automatic transformations using frameworks like those discussed in section~\ref{data:test-sets}.

\begin{table*}[th!]
\small
\centering
\begin{tabular}{@{}lrrr@{}}
\toprule
\textbf{Best Practice \& Implementation} & \textbf{Yes} & \textbf{No} & \textbf{\%} \\ \midrule
\textbf{Make informed evaluation choices and document them}  &  &  &  \\
\hspace{2em} Evaluate on multiple datasets   & 47 & 9 & 83.9 \\
\hspace{2em} Motivate dataset choice(s)   & 21 & 34 & 38.2 \\
\hspace{2em} Motivate metric choice(s)   & 20 & 46 & 30.3 \\
\hspace{2em} Evaluate on non-English language  & 19 & 47 & 28.8 \\
\textbf{Measure specific generation effects} &  &  &  \\
\hspace{2em} Use a combination of metrics from at least two different categories & 36 & 27 & 57.1 \\
\hspace{2em} Avoid claims about overall ``quality'' & 34 & 31 & 52.3 \\
\hspace{2em} Discuss limitations of using the proposed method & 19 & 46 & 29.2 \\
\textbf{Analyze and address issues in the used dataset(s) } &  &  &  \\
\hspace{2em} Discuss or identify issues with the data & 19 & 47 & 28.8 \\
\hspace{2em} Contribute to the data documentation or create it if it does not yet exist & 1 & 58 & 1.7 \\
\hspace{2em} Address these issues and release an updated version & 3 & 10 & 23.1 \\
\hspace{2em} Create targeted evaluation suite(s) & 14 & 52 & 21.2 \\
\hspace{2em} Release evaluation suite or analysis script & 3 & 63 & 4.5 \\
\textbf{Evaluate in a comparable setting } &  &  &  \\
\hspace{2em} Re-train or -implement most appropriate baselines & 40 & 19 & 67.8 \\
\hspace{2em} Re-compute evaluation metrics in a consistent framework & 38 & 22 & 63.3 \\
\textbf{Run a well-documented human evaluation} &  &  &  \\
\hspace{2em} Run a human evaluation to measure important quality aspects & 48 &  18 & 72.7 \\
\hspace{2em} Document the study setup (questions, measurement instruments, etc.) & 40 & 9 & 81.6 \\
\hspace{2em} Document who is participating in the study & 28 & 20 & 58.3 \\
\textbf{Produce robust human evaluation results} &  &  &  \\
\hspace{2em} Estimate the effect size and conduct a power analysis & 0 & 48 & 0.0 \\
\hspace{2em} Run significance test(s) on the results & 12 & 36 & 25.0 \\
\hspace{2em} Conduct an analysis of result validity (agreement, comparison to gold ratings) & 19 & 29 & 39.6 \\
\hspace{2em} Discuss the required rater qualification and background & 10 & 38 & 20.8 \\
\textbf{Document results in model cards} &  &  &  \\
\hspace{2em} Report disaggregated results for subpopulations & 13 & 53 & 19.7 \\
\hspace{2em} Evaluate on non-i.i.d. test set(s) & 14 & 52 & 21.2 \\
\hspace{2em} Analyze the causal effect of modeling choices on outputs with specific properties & 16 & 50 & 24.2 \\
\hspace{2em} Conduct an error analysis and/or demonstrate failures of a model & 15 & 51 & 22.7 \\
\textbf{Release model outputs and annotations} &  &  &  \\
\hspace{2em} Release outputs on the validation set & 1 & 65 & 1.5 \\
\hspace{2em} Release outputs on the test set & 2 & 63 & 3.1 \\
\hspace{2em} Release outputs for non-English dataset(s) & 1 & 25 & 3.8 \\
\hspace{2em} Release human evaluation annotations & 1 & 47 & 2.1 \\ \bottomrule
\end{tabular}
\caption{A condensed view of the recommendations provided in~\cref{sec:recommendations} in a relaxed format for use in our analysis of recent modeling papers (see Appendix~\ref{app:survey} for exact annotation instructions). On the right, we show the number of papers that (do not) follow the recommendations. We also present the percentage of applicable papers that follow each practice. While any one paper should not be expected to follow all the recommendations, a higher overall coverage is highly indicative of a better evaluation process.}
\label{tab:survey}
\end{table*}

To the extent possible, evaluation reports should be framed in causal terms by measuring the response of multiple metrics (or human evaluation) to stimuli to avoid issues with metrics, similar to the CheckList framework by \citet{ribeiro-etal-2020-beyond}. This has the further advantage of setting more realistic user expectations. Taking the example from Figure~\ref{fig:covid}, we could state that 
\say{When the model summarizes news articles from the same source, but with COVID-19 related content, we expect quality drops of 20$\pm$5\% according to \textsc{BLEURT}. N\% of summaries are deemed non-understandable by non-expert raters.}
Evaluation reports should further include improved error analyses, following suggestions by \citet{van-miltenburg-etal-2021-underreporting} and \citet{bender-koller-2020-climbing} who argue for more focus on limitations in addition to aggregated scores. 

Advocating creating evaluation reports does not mean that we should not demonstrate improvements at all, but need to move away from them being the only contribution. Papers should show brittleness and a clear path toward improvements for future work, rather than hiding or being ignorant of existing issues. Another advantage of this framing is that the reliance on large models may dwindle, since work on quantifying shortcomings is equally applicable to smaller models and methods that improve model robustness often work on all model sizes. The explicit set of evaluations that should be run are subject to investigation in future work and may also depend on the claims that are being made.

\section{Every Cloud has a Silver Lining}
\label{sec:survey}

While a survey of challenges and issues will, by definition, paint a rather gloomy picture, there are many positive examples of model evaluations, a couple of which we highlight in this section. To do so, we analyze 66 papers from ACL, INLG, and EMNLP 2021 and the extent to which they already follow our recommendations. Our analysis focuses on whether the different aspects in~\cref{tab:survey} appear in a paper, rather than measuring its extent or quality. That means, for example, that we only identify the presence of a significance test in a human annotation, not judge whether it is the correct test, and that we identify whether any motivation for using a particular dataset exists, not the soundness of the motivation. 
Through this, we aim to capture an upper bound to the existence of the different aspects. We provide additional information on the annotation process and the exact instructions in Appendix~\ref{app:survey}.

\begin{figure}[tb]
    \centering
    \includegraphics[width=0.49\textwidth]{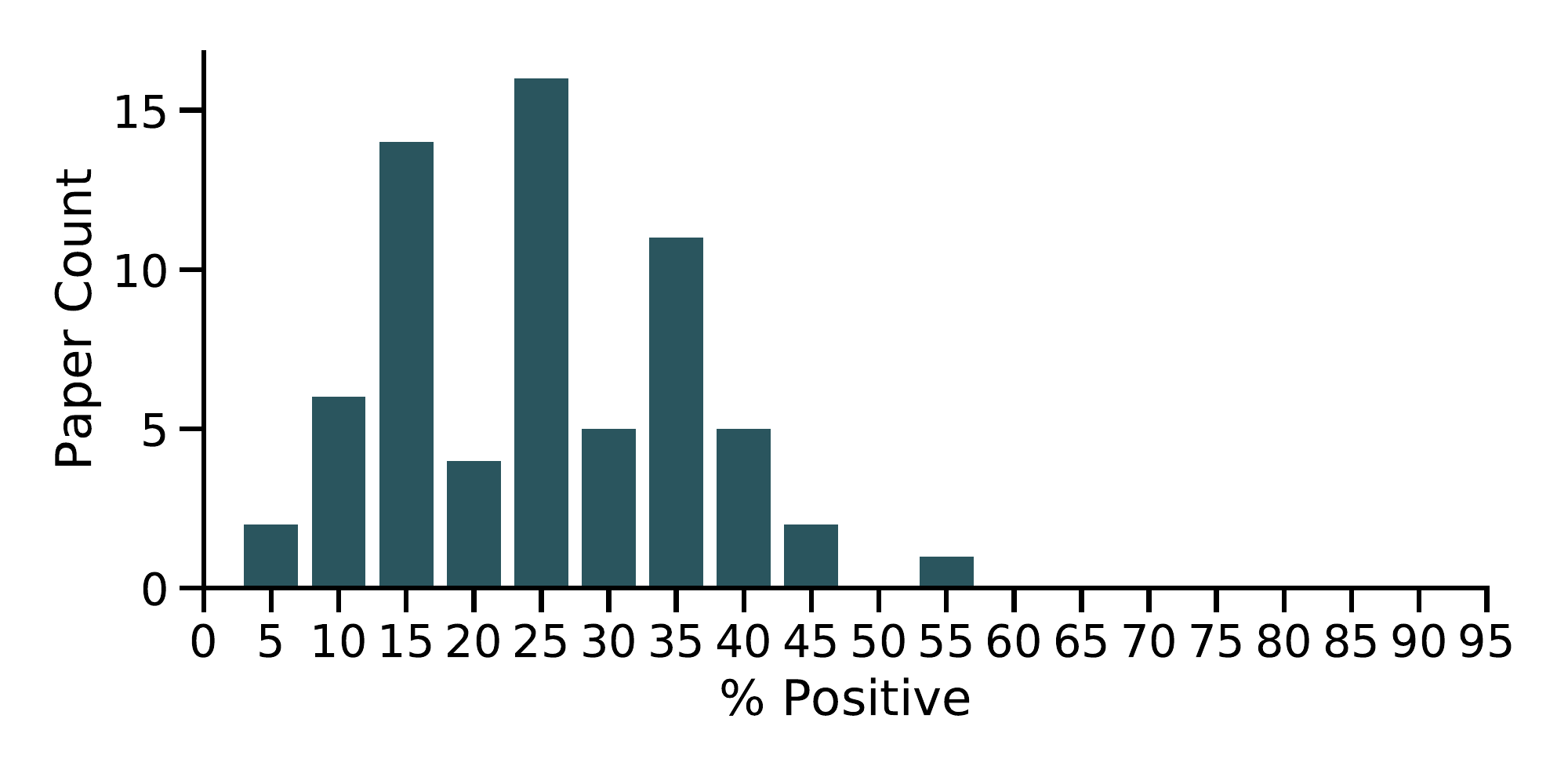}
    \caption{Histogram of the average analysis results per paper. Most papers follow between 15 and 35\% of the suggestions, with much room for improvement.}
    \label{fig:histogram}
    \vspace{-1.5em}
\end{figure}

Overall, we find that 36.7\% of our 2046 judgments were positive, which means that the field has already taken a significant step toward solving the problems pointed out throughout this survey.
Scores for papers ranged from 6.5\% to 58.1\%, with an average of 27.3\% (median 25.8\%, standard deviation of 0.11), demonstrating that there is no consistent standard that is widely applied. We present a histogram of the average scores per paper in~\cref{fig:histogram} and highlight some positive examples of individual judgments when discussing the results. 

Starting with the recommendations pertaining to the basic evaluation setup, the vast majority of papers (84\%) include evaluation results from multiple datasets and reports human evaluation results (73\%).
However, the documentation of the choices that went into the evaluation process and those that relate to the specific claims and motivations is often flawed. Only 38 and 30\% of papers respectively motivate why they chose a particular dataset and metric and half the papers made claims in the abstract pertaining to their system outputs' overall quality when this was not the aspect that was evaluated. 
About 29\% of papers reported results on non-English language, although this result was inflated by machine translation papers included in the analysis. While not explicitly annotated, almost no paper stated that they were working on only English, a practice that has long been criticized~\citep{bender-2009-linguistically}.
Disappointingly, only 29\% discussed the limitation of the proposed method, a finding that corroborates our claim that evaluations are too focused on reporting superior performance rather than fully characterizing system outputs. 
As a positive example, \citet{kim-etal-2021-structure} report negative results on on out-of-distribution performance, encouraging future researchers to work on making their proposed method more robust.

On a positive note, a majority of papers report multiple kinds of metrics. 57\% of papers report metrics from different categories instead of only relying on lexical overlap. In most such cases, the categories were metrics that measure similarity to a reference and diversity among outputs.
However, some also developed metrics to specifically measure what is being claimed. For example, \citet{lyu-etal-2021-encouraging} work on lexical consistency for document-level MT, which they first analyze and then derive a metric from. This metric is subsequently used alongside other metrics to validate their specific claims. 
About 20\% of papers provide additional breakdowns of the results, report on non-i.i.d. test sets, conduct error analyses, or demonstrate a causal effect of input features. 
These are especially helpful when the analysis is motivated by problem-specific needs. For example, \citet{krishna-etal-2021-generating} investigate the generation of doctors' notes from conversations and analyze the performance in the presence of speech recognition errors. 

While 29\% of papers point out issues in the datasets they use or introduce, we found only one paper that contributed to the data documentation, leaving future researchers to rediscover the same issue(s). Moreover, only 3/13 papers that point out issues actually work toward solving them and release updates to the dataset. As discussed above, this is an area where normalizing contributing documentation and releasing updates would have beneficial effects for future work with these datasets. 

Looking closer at the human evaluations, we did not require the definition for each evaluation criterion to be stated, and, therefore, our results look more positive than those by \citet{howcroft-etal-2020-twenty}. We find that 82\% of papers that report human evaluation results also state \textit{what} is being measured, although the documentation of \textit{who} is evaluating is still lacking (58\%). 
However, we did not find a single paper that estimated how many annotations should be collected, and most opted for the ``typical'' 100 data points which, as pointed out above, may be insufficient~\citep{DBLP:journals/csl/LeeGMK21}. Similarly, only 25\% and 39\% of papers assess the annotations and/or the annotators and only 21\% discuss which background knowledge was required to participate in an evaluation. 

An aspect that needs to consistently be improved is the release of data. Even though many papers released datasets or code to reproduce their models, almost none released model outputs or their human evaluation data. This practice can lead to issues when, for example, new papers are not able to compare using the same metrics environment, something that 37\% of papers did not do. Moreover, it can significantly slow down evaluation research due to a lack of data to annotate or human annotations to compare to.

Overall, this analysis shows that there is much room for improvement, but it also shows that we are not starting at zero. While none of the papers reached 100\%, which may be an overly ambitious goal, many reached 40\% or higher, meaning that they already included many of our suggestions.

\section{Discussion}
\label{sec:discussion}

\paragraph{The perfect evaluation is a white whale}

The myriad of evaluation obstacles will not suddenly vanish as we develop metrics that do not suffer from the same shortcomings as human-likeness measures or as we develop better datasets and evaluation suites. Data and evaluations are by design subjective and only reflect a small subset of the space of potential inputs. 
In addition, most critiques and suggestions covered in this paper assume that generated text already exists, but as \citet{dodge-etal-2019-show} demonstrate, compute budget and hyperparameter tuning can also massively confound results. 
The perfect evaluation process is thus not an achievable goal and the question this paper originally set out to answer, \say{How can we fix NLG evaluation?}, does not have a simple answer. However, we pointed many steps toward a \textit{better} evaluation process that will hopefully address these issues. 

This leads to the natural question if it is worth spending all this extra effort on model evaluations when anything we come up with will always remain deeply flawed. Aside from the utilitarian argument that reducing the number of issues are most definitely a beneficial outcome, we pose that creation of evaluation reports should not lead to significant extra effort. There are many research projects focusing on developing evaluation infrastructure, from human evaluation datasheets~\citep{DBLP:journals/corr/abs-2103-09710}, to NLG-specific data cards with interactive collection tools~\citep{mcmillan-major-etal-2021-reusable}, metrics frameworks~\citep{gehrmann-etal-2021-gem}, and APIs that aim to produce replicable human evaluation results~\citep{DBLP:journals/corr/abs-2101-06561}.
Similar to deep learning modeling frameworks, there is a delayed payoff due to their learning curves, but reusing and improving the existing infrastructure will help strengthen evaluations.  

In the end, evaluation practices will only be adopted if reviewers hold model developers accountable to use them. One aim of this work is to document failures in evaluation processes that frequently happen, many of which can be directly addressed or pointed out in reviews. In the future, we also suggest creating model evaluation checklists like those by \citet{rogers-etal-2021-just-think} for responsible data use or \citet{dodge-etal-2019-show} for reporting hyperparameters and compute infrastructure.

\paragraph{Model Interpretability}

This paper omits discussions of analysis methods from the interpretability literature as a part of a model's holistic evaluation process since interpretability may be seen as orthogonal to and is not strictly necessary for model evaluations. However, interpretability methods are a useful tool for evaluations, since they may be used to uncover model shortcuts~\citep{mccoy-etal-2019-right} or to find systematic errors~\citep{DBLP:journals/coling/PopovicN11} that would be described in an evaluation report.\footnote{See \citet{belinkov-glass-2019-analysis} for a recent survey of analysis methods.}
Interpretability tools can also facilitate evaluations. For example, the the Language Interpretability Tool~\citep{tenney-etal-2020-language} can automatically evaluate on subpopulations, and Explainaboard~\citep{liu-etal-2021-explainaboard}, although with limited support for NLG, helps identifying challenging inputs for a model.
Evaluating the interpretability of a model itself as an additional dimension is extrinsic and task-specific~\citep{doshi2017towards} and thus out of scope for this work.

\paragraph{NLG is not ML and also not NLU}

A recent survey by \citet{liao2021are} summarizes evaluation failures across all of machine learning, including computer vision and NLP. While we find overlap with their findings on the data side where most (but not all) criticisms can be applied to other tasks, most of the issues surveyed here go beyond one-size-fits-all machine learning analyses; since outputs are natural language, no equivalent of accuracy or F1-Score exists. This property was focused on by \citet{dale1998towards} who position the evaluation of NLG systems as complementary to NLU systems, pointing out the symmetry of moving from natural language to representations of meaning or structure (NLU) or the other way around (NLG). However, they also discuss evaluation of intermediate steps of an NLG system.
As a consequence of the move toward end-to-end approaches, many NLG tasks have NLU components like the selection of appropriate content that are implicitly evaluated as part of the evaluation of the final generated text. Yet, as \citet{bender-koller-2020-climbing} discuss, these NLU steps require abilities that current models are incapable of acquiring from supervised learning. Evaluating NLG tasks only through the lens of outputs is thus insufficient and we should strife to deliver a more fine-grained breakdown, but it is unclear how to evaluate intermediate steps in current evaluation setups. While separate reasoning steps are starting to be incorporated into current NLG approaches~\citep[e.g.,][]{DBLP:conf/aaai/Puduppully0L19,narayan2021planning}, there has been no consensus for how a planning stage should look like or how to evaluate it and all current evaluation practices are focused on output forms. Nevertheless, approaches that incorporate explicit steps to attribute generated information to sources will be crucial to making progress in the field, and evaluation processes need to reflect these advances~\citep{DBLP:journals/corr/abs-2112-12870}.

\paragraph{The use of models and external evaluation}

A limitation of this work, and evaluation in general, is the focus on intrinsic evaluations and the lack of extrinsic evaluation, and more generally measurement of the external effects of model training and development.
NLG model behaviors may oftentimes be acceptable in some context and undesirable in others.
This is not a problem that can be solved through only intrinsic evaluations since norms are established through language and the cultural background of a person may lead to a different perception of language~\citep{nakayama2011handbook}.
Take for example a summarization system that is run on a subjective article such as a column in a newspaper. A general-purpose summarizer will likely not generate text that states \say{Author X states that Y}, but instead will present opinions such as \say{I don't like math} or \say{Jollof Rice is tasty} as facts. 
This presentation, alongside the anthropomorphic bias of deep learning models~\citep{DBLP:journals/mima/Watson19}, can perpetuate these opinions including harmful stereotypes.
This is a general limitation of NLG models which we are unable to capture using standardized benchmarks alongside intrinsic evaluations.
We also note that few, if any, benchmark currently reports the environmental side-effects of training and serving NLG models~\cite{strubell-etal-2019-energy}.
This means that there are still many considerations required to understand NLG systems that fall beyond the scope of this survey. 
External evaluation may further be more appropriate for interactive systems like dialog systems which are out of scope for this work and which require evaluation considerations such as the distinction between turn- and dialog-level metrics~\citep{DBLP:journals/corr/abs-2201-04723} and which are much more susceptible to antropomorphism~\citep{DBLP:journals/corr/abs-2107-03451}.  

\paragraph{Better metrics will lead to better models}

A common assumption that has been explored with various success in the past is whether it is possible to directly optimize metrics using reinforcement learning instead of the typical cross-entropy objective in NLG tasks. For example, \citet{paulus2018a} demonstrated that we may be able to optimize \textsc{ROUGE} directly and \citet{pasunuru-bansal-2018-multi} explored alternative optimization targets such as maximizing entailment. 
This work assumes that metrics are good proxies for task performance which, as seen in this survey, is demonstrably false. 
In addition, \citet{Choshen2020On} show that improvements from reinforcement learning objectives in machine translation are unrelated to the training signals, but rather a side effect from changes in the model distribution curve. 
They even find this to be true when the reward signal is semantic similarity instead of \textsc{BLEU}~\citep{wieting-etal-2019-beyond}. 
However, more recent work demonstrates that, in machine translation, optimizing toward newer learned metrics like BLEURT does not suffer from this issue, leading to significant model improvements~\citep{DBLP:journals/corr/abs-2104-07541}. 
Developing better metrics thus provides an exciting opportunity to close the circle between metrics and models, especially if we can optimize toward multiple metrics which measure disjoint quality aspects.
However, this advance relies on our recommendation for evolving metrics and the embrace of deprecation.







\section{Conclusion}

We surveyed challenges in NLG evaluation from the perspective of automatic metrics, human evaluation, and datasets. Our findings reveal that, while much progress is being made, the evaluation process currently applied to most models is not sustainable. Models have improved to the point where differences between them are unlikely to be spotted based on surface-level phenomena and careful annotation process are required to characterize their output quality and distinguish between them. 
Moreover, we discussed issues with popular NLG datasets that further conflate evaluation results.

In addition to pointing to worthwhile evaluation-related research directions, we suggest a series of actionable improvements that model developers can follow that will have a positive long-term impact while also improving their model evaluations. 
Notably, we argue for evaluation reports that focus on a causal framing of limitations of models with the goal to eventually be able to provide performance guarantees for a wide set of potential deployment scenarios. 
We show in an analysis of 66 recent NLG papers, that many of the suggestions are partially followed already, but that there is no consistent standard which evaluation aspects are required of researchers.
When discussing the limitations of our suggestions, we note that their implementation will require changes to the peer review system to hold model developers accountable to follow the suggested best practices. 
\section*{Acknowledgements}

We are grateful to Mirella Lapata, Ankur Parikh, Dipanjan Das, and Slav Petrov, who have provided comments on earlier versions of this paper. The content of this work was additionally discussed with many others, including Matthew Lamm, Vitaly Nikolaev, and many of the participants in the GEM benchmark. Without those discussions, the subsections and discussion points would look very differently and we thank everyone who participated in the discussions. 

\bibliographystyle{acl_natbib}
\bibliography{anthology,acl}

\appendix
\newpage
\section{Surveying recent ACL, INLG, and EMNLP papers}
\label{app:survey}

Here we describe the annotation instructions for our analysis of 66 ACL, EMNLP, and INLG papers from 2021. The instructions were defined such that results are an upper bound to the criteria. We avoid judging quality of a particular evaluation aspect and instead only annotate its presence.

\subsection{Paper selection}

The analyzed papers were selected from the proceedings of the three conferences. A paper was selected if it included in its title any reference to working on a generation problem. To avoid an over-emphasis on machine translation in the analysis, we did not specifically select translation papers unless their title was related to generation as a whole. As a result, about 10--15 translation papers were part of the analysis which we considered an appropriate amount. 
The selected papers were subsequently filtered if they did not provide modeling results, e.g., because it was an analysis-focused paper or if the modeling task was not a generation task covered by our definition in \cref{sec:background}. 

\subsection{Instructions}

\vspace{1em}

\noindent \textbf{Make informed evaluation choices and document them}    

\begin{itemize}  \setlength\itemsep{0em}
\item Evaluate on multiple datasets: Select yes if the paper reports results on more than one dataset. Select N/A if the paper explicitly states that there is only one dataset available for the addressed task.
\item Motivate dataset choice(s): Select yes if the paper states why each particular dataset was chosen. If the only reasoning is that previous work uses it, select no. If the paper introduces a dataset, select N/A.
\item Motivate metric choice(s): Select yes if the paper states why each particular metric was chosen. If the only reasoning is that previous work uses it, select no.
\item Evaluate on non-English language: If at least one of the evaluated datasets includes non-English language, select yes.    
\end{itemize}

\noindent \textbf{Measure specific generation effects}   

\begin{itemize}  \setlength\itemsep{0em}
\item Use a combination of metrics from at least two different categories: Select yes, if the automatic evaluation results include at least two metrics from different families (e.g., one QA-based one and one lexical one). Reporting ROUGE and BLEU would not count while ROUGE and BLEURT would.   
\item Avoid claims about overall ``quality'': Select no if \textbf{the abstract} of the paper reports improvements generally and not in terms of specific generation aspects (e.g., ``we outperform baselines'')   
\item Discuss limitations of using the proposed method: Select yes, if there is at least one paragraph dedicated to the limitations of the proposed method in the results or discussion section or as its own section.   
\end{itemize}

\noindent \textbf{Analyze and address issues in the used dataset(s)}   

\begin{itemize}  \setlength\itemsep{0em}
\item Discuss or identify issues with the data: Select yes, if there is at least a mention of problematic artefacts with the data or what or who it represents.   
\item Contribute to the data documentation or create it if it does not yet exist: Select yes, if the paper is accompanied by a data card or if there is a mention that original documentation was updated.
\item Address these issues and release an updated version: Select yes, if the paper is accompanied by a release of updated data or points to a loader that retrieves the updated dataset. If the paper introduces a dataset, select N/A.
\item Create targeted evaluation suite(s): Select yes, if the paper describes the creation of a fine-grained breakdown of subpopulations \textbf{or} multiple training or test splits.
\item Release evaluation suite or analysis script: Select yes, if the resources in the previous points were released in the form of data or code.
\end{itemize}

\noindent \textbf{Evaluate in a comparable setting }   

\begin{itemize}  \setlength\itemsep{0em}
\item Re-train or -implement most appropriate baselines: Select yes, if the paper explicitly mentions that it trains or implements baselines from prior papers.
\item Re-compute evaluation metrics in a consistent framework: Select yes, if \textbf{all} the reported scores were computed by the authors or by another centralized framework (e.g., through upload to a leaderboard). If only a subset was recomputed, select no.
\end{itemize}

Select N/A for both questions above if a new dataset was introduced and the only one evaluated in the paper.

\noindent \textbf{Run a well-documented human evaluation}   

\begin{itemize}  \setlength\itemsep{0em}
\item Run a human evaluation to measure important quality aspects: Select yes, if a human evaluation of any kind was conducted.
\item Document the study setup (questions, measurement instruments, etc.): Select yes, if, at the minimum, the specific questions and the way that participants answer them are reported. 
\item Document who is participating in the study: Select yes, if, at the minimum, the annotation platform used and the number of participants are stated. 
\end{itemize}

\noindent \textbf{Produce robust human evaluation results}   

\begin{itemize}  \setlength\itemsep{0em}
\item Estimate the effect size and conduct a power analysis: Select yes, if any effect size estimate or power analysis is mentioned (we assume that not mentioning it implies it absence).  
\item Run significance test(s) on the results: Select yes, if the human annotation results are accompanied by a statistical significance test.
\item Conduct an analysis of result validity (agreement, comparison to gold ratings): Select yes, if there is any kind of analysis of the quality of the human annotations themselves.
\item Discuss the required rater qualification and background: Select yes, if the required knowledge of raters is discussed and compared to the qualifications selected for in the study. 
\end{itemize}

\noindent \textbf{Document results in model cards}   

\begin{itemize}  \setlength\itemsep{0em}
\item Report disaggregated results for subpopulations: Select yes, if the paper reports fine-grained results on subsets of the test set(s) (note that the paper does not need to introduce these breakdowns as in the point above).
\item Evaluate on non-i.i.d. test set(s): Select yes, if there is an evaluation on a non-i.i.d. test set. If the paper does not specifically mention this fact, select no (i.e., if the used dataset has such a test set but this is not mentioned).
\item Analyze the causal effect of modeling choices on outputs with specific properties: Select yes, if the results include a breakdown that allow for insights of the form ``if input has feature X, model output has Y''. An ablation study counts as a yes, \textbf{if} the ablation focuses on feature representations (i.e. what data a model sees), but not if the ablation is on model architecture choices.
\item Conduct an error analysis and/or demonstrate failures of a model: Select yes, if there is any kind of error analysis or qualitative samples of where the model fails.
\end{itemize}

\noindent \textbf{Release model outputs and annotations}   

In this section, select yes, if the paper is accompanied by data releases that include the following.

\begin{itemize}  \setlength\itemsep{0em}
\item Release outputs on the validation set 
\item Release outputs on the test set   
\item Release outputs for non-English dataset(s): Select N/A if the paper does not include evaluation on any non-English data.
\item Release human evaluation annotations 
\end{itemize}

\subsection{Limitations}

There are a few limitation of this setup. (1) Due to the phrasing as recall-oriented prompts, nuanced errors pointed out in earlier sections are implicitly ignored.  For example, ``Document the study setup'' is marked as positive even if the exact definition of each measurement category is not provided. The lack of providing a definition was identified as a source of confusion by \citet{howcroft-etal-2020-twenty}.
In other cases, our prompts may not be covering all possibilities. For example, a study that releases not an improved version of a corpus, but instead a tailored pretraining set would not count as ``Address dataset issues and release an updated version''.  
(2) Each paper is only annotated by one co-author of this survey. This means that there could be misunderstandings of the different dimensions. We tried to address this problem by refining definitions when unclear points arose and by discussing the definitions before starting the annotation which led to the instructions above. Nevertheless, the exact percentage results may differ from the ground-truth by a few points and we thus consider only the overall trends.

\end{document}